\documentclass[runningheads]{llncs}
\usepackage{eccv}
\usepackage{eccvabbrv}
\usepackage{graphicx}
\usepackage{booktabs}
\usepackage[accsupp]{axessibility}
\usepackage[most]{tcolorbox}
\usepackage{hyperref}
\usepackage{orcidlink}
\begin{document}

\title{Text-Conditioned Background Generation for Editable Multi-Layer Documents} 

\titlerunning{Text-Conditioned Background Generation}

\author{Taewon Kang\inst{1}\thanks{Work done while first author was an Adobe Research intern and then continued as a collaborative effort with UMD.}\orcidlink{0009-0004-8534-6314} \and
Joseph K J\inst{2}\orcidlink{0000-0003-1168-1609} \and
Chris Tensmeyer\inst{2} \orcidlink{0000-0002-0761-3243} \and
Jihyung Kil\inst{2} \orcidlink{0009-0005-7044-2781} \and
Wanrong Zhu\inst{2} \orcidlink{0009-0005-3448-0078} \and
Ming C. Lin\inst{1}\orcidlink{0000-0003-3736-6949} \and
Vlad I. Morariu\inst{2}\orcidlink{0000-0001-7937-7748}}
\authorrunning{T.~Kang et al.}

\institute{University of Maryland at College Park, United States \and
Adobe Research, United States \\
\email{taewon@umd.edu}, \email{\{josephkj,tensmeye,jkil,wzhu\}@adobe.com}, \email{lin@umd.edu}, \email{morariu@adobe.com}}

\maketitle

\begin{abstract}
  We present a framework for document-centric background generation with multi-page editing and thematic continuity. To ensure text regions remain readable, we employ a \emph{latent masking} formulation that softly attenuates updates in the diffusion space, inspired by smooth barrier functions in physics and numerical optimization. In addition, we introduce \emph{Automated Readability Optimization (ARO)}, which automatically places semi-transparent, rounded backing shapes behind text regions. ARO determines the minimal opacity needed to satisfy perceptual contrast standards (WCAG 2.2) relative to the underlying background, ensuring readability while maintaining aesthetic harmony without human intervention. Multi-page consistency is maintained through a summarization-and-instruction process, where each page is distilled into a compact representation that recursively guides subsequent generations. This design reflects how humans build continuity by retaining prior context, ensuring that visual motifs evolve coherently across an entire document. Our method further treats a document as a structured composition in which text, figures, and backgrounds are preserved or regenerated as separate layers, allowing targeted background editing without compromising readability. Finally, user-provided prompts allow stylistic adjustments in color and texture, balancing automated consistency with flexible customization. Our training-free framework produces visually coherent, text-preserving, and thematically aligned documents, bridging generative modeling with natural design workflows.
\end{abstract}
\vspace*{-3em}
\section{Introduction}
\vspace*{-1em}

\begin{figure*}[t]
    \centering
    \includegraphics[width=0.925\linewidth]{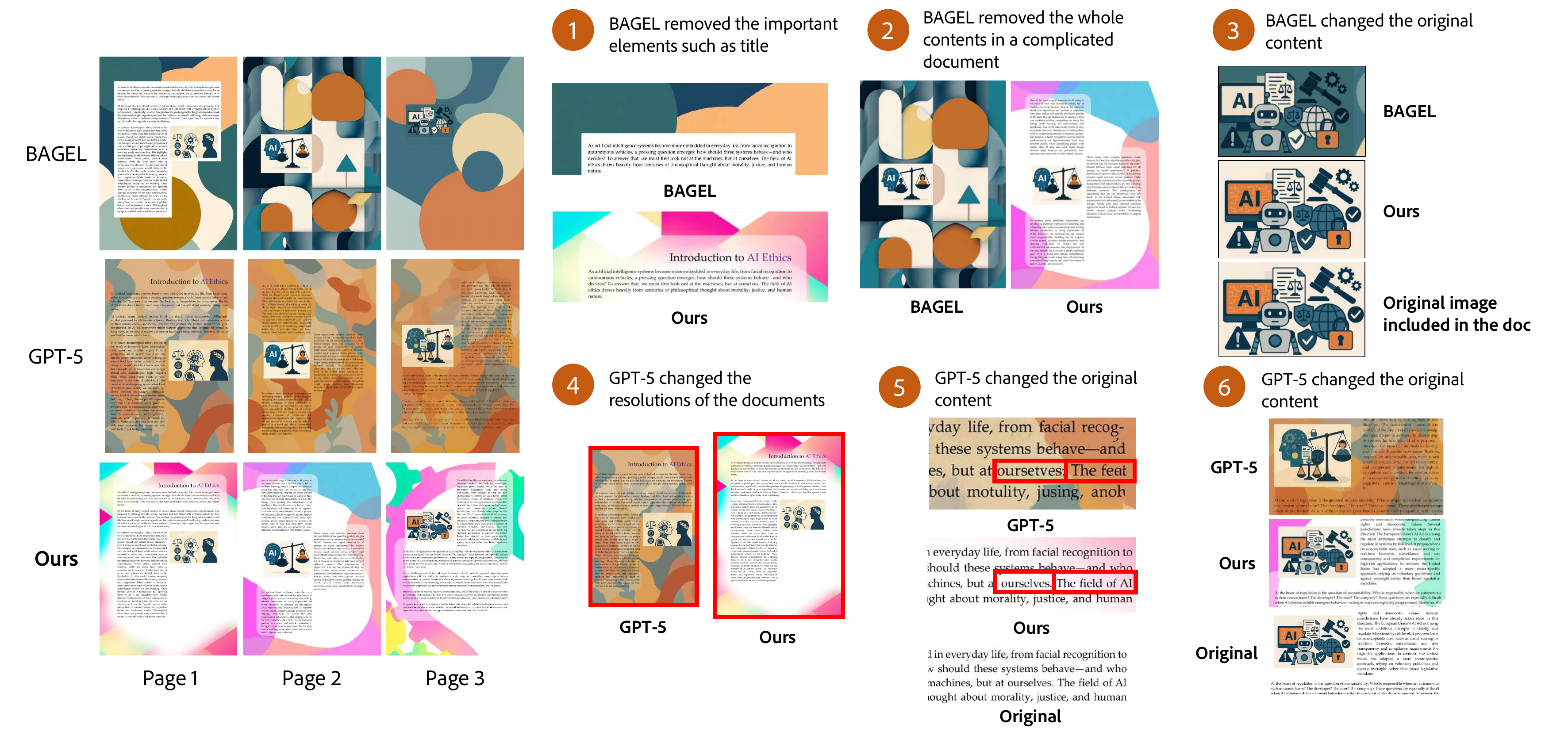}
    \vspace*{-1em}
    \caption{\textbf{Comparison with existing diffusion methods.} Baseline diffusion models overwrite or alter the original document: removing titles and figures ((1), (2)), modifying semantic content ((3),(5),(6)), and even changing resolution ((4)). In contrast, our method preserves all foreground elements (text + images), while generating visually coherent, multi-page backgrounds aligned with the document content.}
    \vspace*{-3em}
    \label{fig:teasure}
\end{figure*}

Designing and editing complex documents that interleave text and images—such as academic reports, educational materials, or presentation slides—remains a longstanding challenge in generative modeling. While diffusion models have made significant progress in text-to-image synthesis and style transfer, they are typically optimized for producing standalone images rather than for documents. Applying diffusion techniques to documents requires not only producing visually coherent backgrounds but also ensuring text readability, layout fidelity, and consistency across multiple pages. Among these, background generation is particularly critical because visually salient backgrounds can interfere with readability if not handled properly, yet existing systems often under-emphasize this aspect, resulting in visual artifacts, results appearing disjointed across pages, or inconsistent with intended designs~\cite{deng2025emerging,openai2024gpt4o,openai2025introducinggpt5}.  

A particularly important aspect of this task is the ability to perform fine-grained background editing while maintaining global coherence.  Designers may wish to adjust motifs, alter stylistic elements, or refine individual pages—without disrupting the overall flow of the document. Standard diffusion pipelines, however, often struggle to reconcile localized edits with global consistency, leading to cross-page inconsistencies~\cite{deng2025emerging,openai2024gpt4o,openai2025introducinggpt5}, thereby hindering integration of generative models into real-world document workflows.

To address these gaps, we propose a document background design framework that adapts diffusion for layered, multi-page documents.
Our system focuses on the specific challenge of background generation while preserving readability and layout. Our approach treats a document not as a flat image but as a structured multi-layer composition, where text, figures, and backgrounds can be independently preserved or regenerated. This layered representation enables selective modification of backgrounds while safeguarding textual content through layout-aware conditioning. At the same time, our framework emphasizes the novelty of multi-page consistency: each page is summarized into a compact representation, and this summary recursively~\cite{kang2025action2dialogue} guides subsequent background generation through concise design instructions. This process ensures that visual motifs evolve coherently across an entire document, rather than drifting page by page (Fig~\ref{fig:teasure}). 

A central component of this framework is a strategy called \emph{latent masking}, which protects foreground regions such as text and figures during background generation. Salient background objects often obscure document readability if left unchecked. A naive solution is to erase content using binary masks, but such hard removal introduces sharp boundaries and produces visually jarring results. Instead, our approach begins from the key idea of \textbf{purposefully weakening background generation} in foreground areas to ensure their preservation. We implement this through a soft, layout-aware attenuation mask applied in latent space, which reduces diffusion updates in regions containing text while allowing the background to evolve naturally around them. Without applying hard constraints to create discontinuities, we model masking as a smooth attenuation field in latent space, analogous to diffusion barriers in physics or weighting functions in numerical optimization. This formulation allows the background to evolve naturally around protected regions while minimizing artifacts at the boundaries, producing visually rich yet balanced results -- supporting iterative refinement across pages.

Beyond masking, we introduce \emph{Automated Readability Optimization (ARO)} to guarantee text legibility. Unlike prior systems that place uniform opaque patches behind text, ARO automatically determines~\cite{kang2021multiple} the minimal opacity of semi-transparent, rounded-corner backing shapes needed to satisfy perceptual contrast standards (WCAG~2.2). By analyzing the luminance distribution of the background and blending it with the overlay color, ARO computes an opacity value $\alpha$ that achieves a target contrast ratio across a specified coverage fraction of pixels. This ensures that inserted shapes remain visually harmonious with the background, while meeting accessibility-driven readability requirements without human intervention.

Our method also incorporates semantic grounding to align visuals with document themes. Document-level summaries guide the background generation process, aligning motifs with subject matter (e.g., fairness in AI ethics, molecular structures in life sciences). In addition, natural language prompts allow stylistic adjustments, enabling users to influence color palettes, textures, and other aesthetic choices. This combination of automated summarization with optional user constraints supports both consistent automation and flexible customization.

\begin{figure*}[t]
    \centering
        \includegraphics[width=\linewidth]{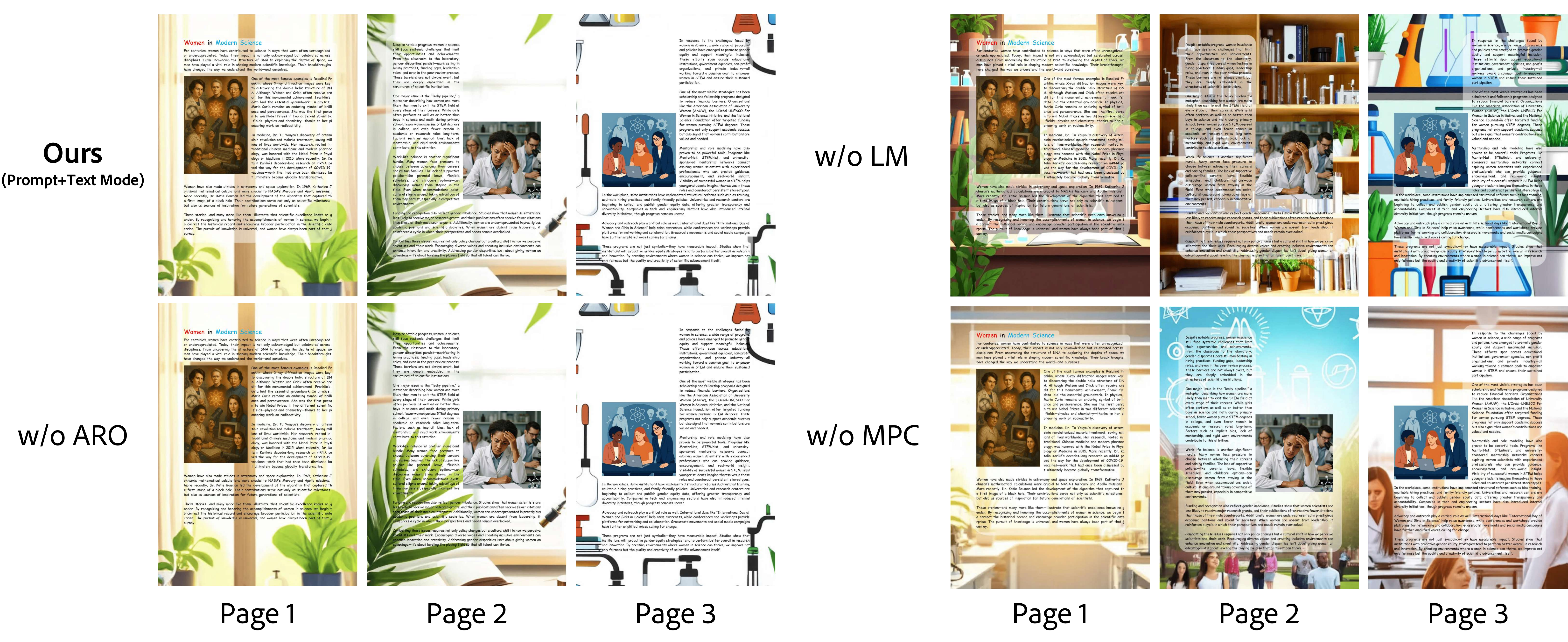}
        \vspace*{-2em}
    \caption{ \textbf{Ablation on our document-aware background generation.} \textbf{Ours} (left) preserves readability and maintains consistent visual themes across pages. \textbf{w/o LM} (no latent masking) allows background objects to intrude into foreground text and images; ARO cannot recover readability due to intrusion. \textbf{w/o ARO} (no readability optimization) keeps the theme, but text becomes harder to read due to insufficient contrast. \textbf{w/o MPC} (no multi-page consistency) produces different styles on each page, losing cross-page visual coherence. }
    \label{fig:ablation_latentmasking_aro}
    \vspace*{-2.5em}
\end{figure*}

This work highlights how diffusion models can be adapted from pure image synthesis to structured, multi-page document background generation. By emphasizing readability, consistency, and controllability, our framework bridges generative modeling with the practical needs of document editing.
Our contributions are three-folded:
\begin{enumerate}
    \item \textbf{Latent Masking for Foreground-aware Backgrounds.} We introduce a layout-aware attenuation in latent space that reduces diffusion updates on text and figure regions while operating in a layer-separable document representation, yielding natural, non-intrusive backgrounds with preserved readability. Instead of enforcing hard binary constraints, our formulation models masking as a smooth attenuation field to support iterative refinement and stylistic consistency across pages with alignment between visual design and document content.
    \item \textbf{Automated Readability Optimization (ARO).} We propose a contrast-driven, WCAG-guided method that computes the \emph{minimal} opacity of semi-transparent rounded backings per text box for coverage-aware linear-light computation, guaranteeing legibility without manual tuning (see Fig.~\ref{fig:ablation_latentmasking_aro}).
    \item \textbf{LLM-based Multi-Page Consistency.} We employ a two-stage pipeline with a \emph{Summarization Model} (to distill each page into a compact representation) and an \emph{Instruction Generation Model} (to convert summaries into concise background design prompts), recursively carrying context across pages to maintain consistent motifs.
\end{enumerate}

\section{Related Work}
\vspace*{-1em}
\subsection{Document Background and Poster Generation}
\vspace*{-1em}
Recent works have begun to explore the integration of generative models into background~\cite{eshratifar2024salient,li2023planning,inoue2024opencole,wang2025designdiffusion,huang2024layerdiff,dalva2024layerfusion}, layout~\cite{li2023relation,hu2024ella,luo2024layoutllm,yang2024mastering,weng2024desigen,zhang2025creatidesign,wu2025hybrid} and poster creation~\cite{chen2025posta,zhang2025creatiposter,peng2025bizgen}.
BAGEL~\cite{deng2025emerging} enables intuitive text-to-design and design editing through natural language prompts, lowering the entry barrier for non-expert users. A notable advantage of BAGEL is its strong consistency in interactive editing: once an initial generation is produced, users can reliably make localized modifications to selected elements while preserving the rest of the layout. This makes BAGEL highly effective for poster-style editing. However, BAGEL struggles when applied to complex documents that interleave dense text, figures, and tables across multiple pages, where background synthesis often interferes with content fidelity and fails to maintain cross-page coherence. To address aesthetics, POSTA~\cite{chen2025posta} proposes a modular framework that combines multimodal large language models and diffusion models to generate artistic posters. While this achieves high text accuracy and visually compelling backgrounds, it still requires explicit human intervention through prompt design and planning, restricting full automation. CreatiPoster~\cite{zhang2025creatiposter} further improves fidelity by supporting editable, multi-layer compositions, surpassing existing commercial systems in accuracy and asset handling. Nonetheless, it also depends heavily on human-provided instructions and assets, thereby limiting scalability to large document collections. Across these approaches, the central limitation is a reliance on prompt-heavy or manual workflows, along with weak protection of critical document content such as tables, headers, or densely populated text blocks.

\vspace*{-1em}
\subsection{Diffusion for Design and Layout}
\vspace*{-1em}
Diffusion-based design frameworks such as BAGEL~\cite{deng2025emerging} enable iterative refinement of visual elements once an initial layout is produced, making them effective for poster-like media. However, when applied to documents that interleave dense text and embedded figures, two challenges emerge: (1) background synthesis frequently intrudes into foreground regions, harming readability~\cite{scharff2000discriminability,scharff2003contrast,leykin2004automatic,zuffi2007human,zuffi2009understanding}, and (2) maintaining page-to-page stylistic coherence becomes difficult. Prior work on text-aware generation tackles complementary goals: TextDiffuser and TextDiffuser-2~\cite{chen2023textdiffuser,chen2024textdiffuser} generate legible text using explicit position conditioning, and SAWNA~\cite{10.1145/3721250.3743023} preserves empty negative space by injecting nonreactive noise. Yet these methods assume either controllable text rendering or blank regions; they do not protect existing foreground content in complex document layouts. Classical readability studies~\cite{scharff2000discriminability} further show that when textured backgrounds interact with text, legibility decreases and can only be restored by global masking or frequency filtering. Beyond layout handling, most diffusion-based editing methods focus on quality maximization rather than content preservation. Mask-guided editing~\cite{avrahami2022blended,couairon2022diffedit,lugmayr2022repaint,ju2024brushnet} restricts edits spatially, but text can still be overwritten if masks overlap. Attention-control models~\cite{hertz2022prompt,cao2023masactrl,chefer2023attend} sharpen generation by reinforcing semantics, whereas layered approaches~\cite{huang2024layerdiff,li2023layerdiffusion,zhang2024transparent} assume clean separable layers—an unrealistic assumption for dense document layouts. Spatial or training-free layout control~\cite{sun2024spatial,Chen_2024_WACV,Xie_2023_ICCV} relies on strong external signals and does not inherently prevent degradation of embedded text. As recent surveys note~\cite{10884879}, diffusion research overwhelmingly pursues sharper and more detailed outputs. In contrast, document-centric editing requires a fundamentally different objective: \textbf{to preserve strict text fidelity, the model must sometimes do \emph{less}, not more}. Our latent masking softly attenuates diffusion updates in sensitive regions, analogous to smooth barrier functions in control and optimization~\cite{7782377,Rabiee2023SoftMinimumBF,Wang2022EnforcingHC,4178067}, while our Automated Readability Optimization (ARO) module enforces WCAG contrast standards~\cite{wcag22_w3c2025}. Together, attenuation in latent space and explicit contrast optimization provide principled protection of readability — often overlooked by prior works.

\vspace*{-1em}
\subsection{Interactive Document Editing}
\vspace*{-1em}
A parallel line of research emphasizes interactivity in editing.
Existing frameworks such as POSTA~\cite{chen2025posta}, CreatiPoster~\cite{zhang2025creatiposter}, and InstructPix2Pix~\cite{brooks2023instructpix2pix} adopt human-in-the-loop designs, relying on iterative prompt adjustments, localized editing, asset uploads, or layout preferences to guide the final result. While this improves user control, it also shifts the burden onto the designer, limiting efficiency for multi-page or large-scale editing tasks. More broadly, interactivity is not limited to prompt-based interfaces: conventional authoring tools such as PowerPoint also support interactive document editing. Beyond poster-oriented systems, controllable layout-to-image frameworks such as GLIGEN~\cite{li2023gligen}, LayoutDiffusion~\cite{zheng2023layoutdiffusion}, and HiCo~\cite{ma2024hico} extend interactivity by enabling grounded or hierarchical control over generation. However, these approaches still rely heavily on iterative user intervention and do not fully address the unique constraints of document-centric editing. Generative systems must therefore balance the strengths of automation with the flexibility of iterative human refinement. Moreover, systems like GPT-4o~\cite{openai2024gpt4o} and GPT-5~\cite{openai2025introducinggpt5}, when applied to document backgrounds, alter not only the visual layer but also the textual content itself, rendering them unsuitable for editing use cases. These observations underscore the gap between artistic generation and document-centric editing: robust systems must allow users to modify or regenerate backgrounds interactively while preserving strict text fidelity. This motivates methods that combine structured summarization, memory-driven instruction generation, and latent masking, thereby ensuring automation and interactivity without sacrificing robustness.

\vspace*{-1em}
\section{Method}

\begin{figure*}[t]
    \centering
    \includegraphics[width=0.9\linewidth]{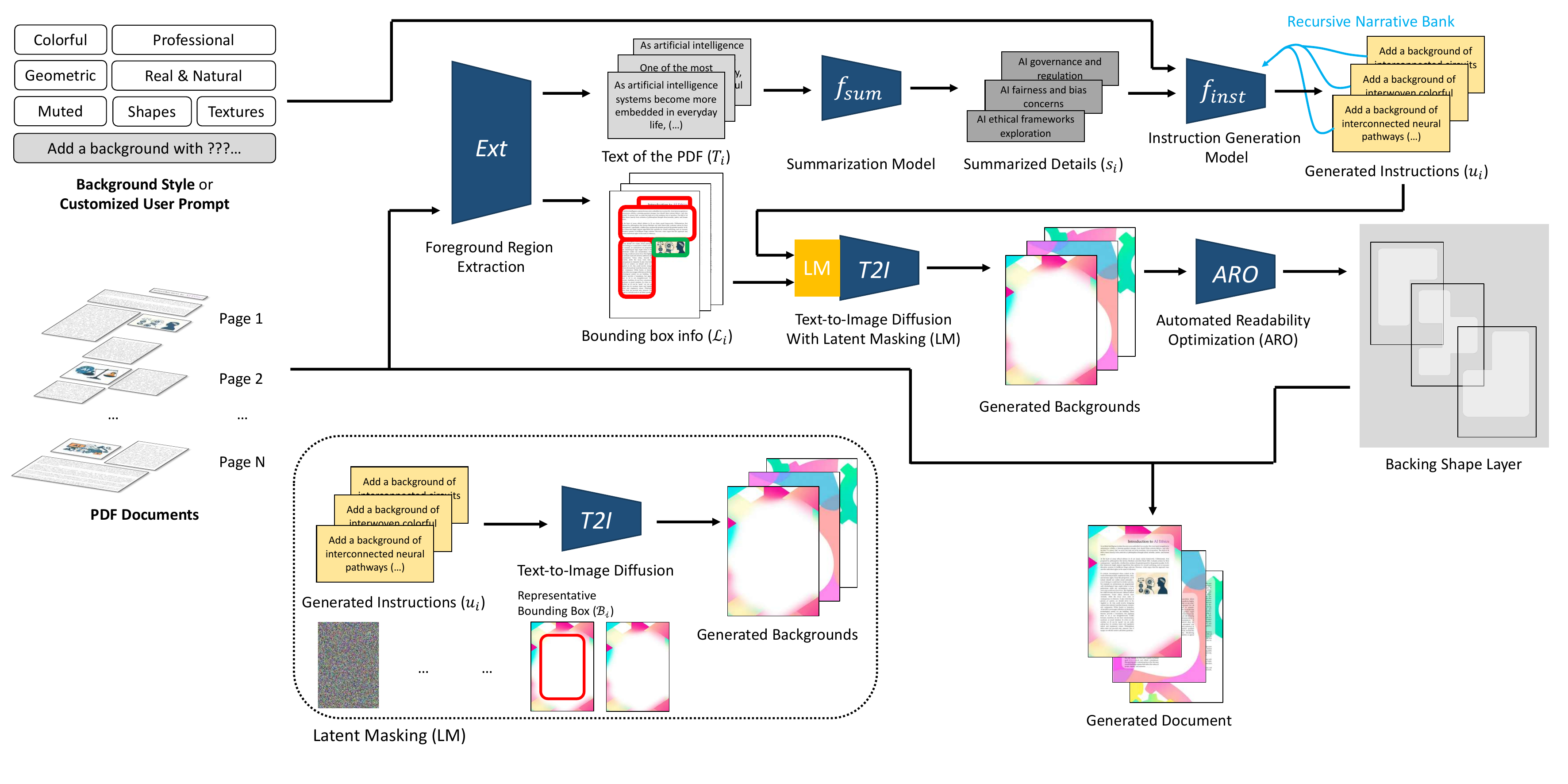}
    \vspace*{-1em}
    \caption{
    \textbf{Overview of our document-centric background generation framework.} Given structured document pages (e.g., PDF, slides), we first perform \emph{Foreground Region Extraction} to obtain page-level text $T_i$ and bounding box information $\mathcal{L}_i$, while selecting representative regions $\mathcal{B}_i$ for latent masking. The \emph{Summarization Model} compresses verbose page text $T_i$ into a compact semantic label $s_i$, which is transformed into generation instructions $u_i$ by the \emph{Instruction Model} with multi-page continuity enforced by the Recursive Narrative Bank (RNB). Backgrounds are then synthesized by a text-to-image diffusion model, guided by (i) \emph{Latent Masking (LM)} using $\mathcal{B}_i$ to preserve foreground readability, and (ii) \emph{Automated Readability Optimization (ARO)} which adaptively places semi-transparent backing shapes behind all text regions $\mathcal{L}_i$ to satisfy WCAG contrast requirements. The resulting backgrounds are composited with the document foreground, yielding coherent, readable, and visually consistent multi-page documents.}
    \label{fig:method_figure_doc}
    \vspace*{-2em}
\end{figure*}

\begin{figure*}[t]
    \centering
    \includegraphics[width=\linewidth]
    {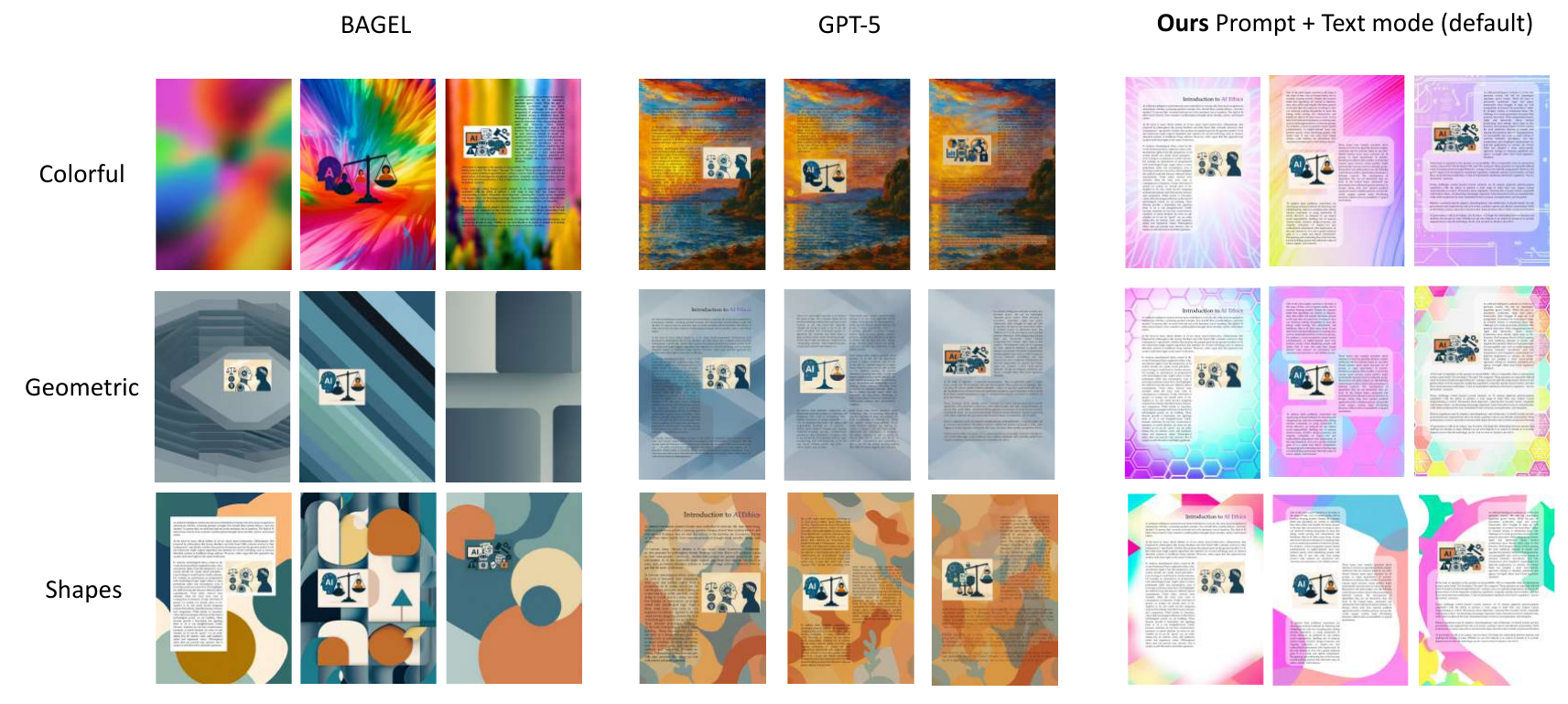}
    \vspace*{-1.5em}
    \caption{Representative qualitative comparison on academic-style \textbf{PDFs} (A4). Rows correspond to style conditions (Colorful, Geometric, Muted, Professional, Real \& Natural, Shapes, Textures).
    See more results in the supplementary materials.
    }
    \label{fig:results_pdfs_representative}
\end{figure*}

\begin{figure*}[t]
    \centering
    \includegraphics[width=\linewidth]{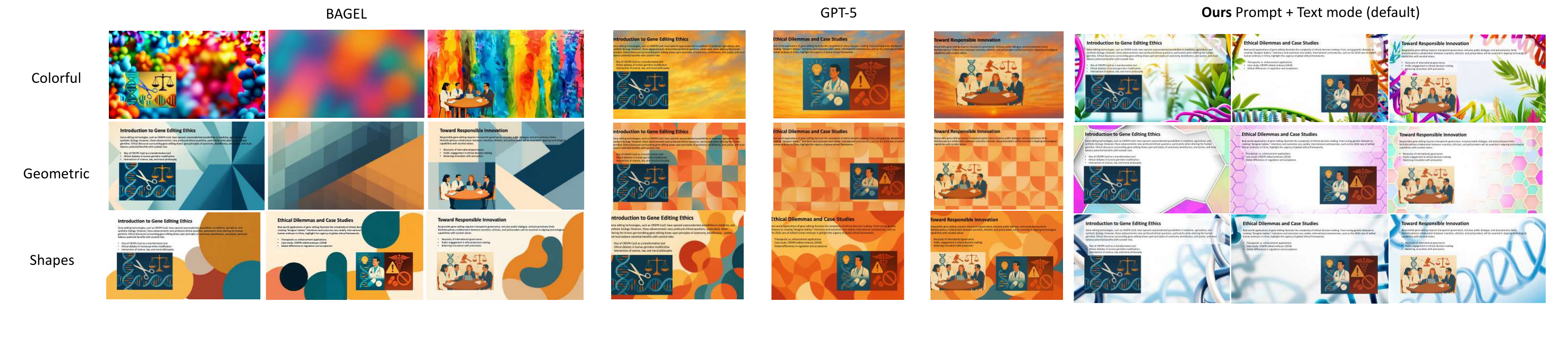}
    \vspace*{-2em}
    \caption{Representative qualitative comparison on academic-style \textbf{slides} (16:9). Rows correspond to style conditions (Colorful, Geometric, Muted, Professional, Real \& Natural, Shapes, Textures).
    See more results in the supplementary materials.
    }
    \label{fig:results_slides_representative}
\end{figure*}

\subsection{Foreground Region Extraction}
Before summarization, we identify representative text regions that define the document foreground. 
For each page $i$, we detect text lines 
$\mathcal{L}_i=\{(\mathbf{b}_{i,\ell},t_{i,\ell})\}$ with bounding boxes 
$\mathbf{b}_{i,\ell}=(x_0,y_0,x_1,y_1)$. Paragraphs are formed by grouping adjacent lines based on 
consistent left margins and bounded vertical gaps. Each paragraph $p$ then receives a box 
$\mathbf{b}_p$ obtained as the union of its constituent line boxes.  

To prevent merging across large figures, detected image zones $\mathcal{I}_i$ partition the page 
vertically into top, side, and bottom groups. Within each group, paragraphs are merged into 
column-like regions if their horizontal overlap  
\begin{equation}
\mathrm{overlap}_x(\mathbf{b}_p,\mathbf{b}_r) \;=\; 
\frac{\max\{0,\, \min(x_1^p,x_1^r)-\max(x_0^p,x_0^r)\}}
     {\max\{1,\,\min(x_1^p-x_0^p,\;x_1^r-x_0^r)\}}
\end{equation}
exceeds a threshold $\eta_x$, and their vertical gap is within tolerance. Finally, an NMS-like suppression removes redundant overlaps: a candidate region $p$ is discarded 
if it is largely contained in a larger region $q$, i.e.,
\begin{equation}
\frac{\mathrm{area}(p\cap q)}{\min\{\mathrm{area}(p),\mathrm{area}(q)\}} \ge \tau_{\mathrm{cont}}
\quad\text{or}\quad 
\mathrm{IoU}(p,q)\ge\tau_{\mathrm{iou}}.
\end{equation}
The surviving set $\mathcal{R}_i=\{(\mathbf{b}_r,t_r)\}$ yields representative bounding boxes 
$\mathcal{B}_i=\{\mathbf{b}_r\}$, while the full-page text is
\vspace*{-0.5em}
\begin{equation}
T_i = \mathrm{concat}\big(\{t_{i,\ell} \mid (\mathbf{b}_{i,\ell},t_{i,\ell})\in \mathcal{L}_i\}\big).
\end{equation}

\subsection{Summarization Model}
To ground visual design in document content, we first extract raw text $T_i$ from each page $i$ of a \emph{structured document instance} (e.g., PDF page, slide canvas, or equivalent container), abstracting away from any specific file format. Document pages often contain multiple paragraphs with dozens of sentences, making $T_i$ verbose and noisy for visual grounding. Directly feeding such heavy text into subsequent modules introduces two problems: (i) generation of overly complex or conflicting background prompts, and (ii) loss of robustness due to long, detailed paragraphs across many densely populated pages.  We map each $T_i$ into a compact semantic label $s_i$: 
\begin{equation} 
    s_i = f_{\text{sum}}(T_i), 
\end{equation} 
where $f_{\text{sum}}$ outputs a short phrase (five words or fewer) capturing the dominant visual theme of the page. This plays a dual role: compressing dense text into a concise representation for visual grounding, and providing semantic cues for background generation. Foreground preservation itself is handled separately using the representative bounding boxes $\mathcal{B}_i$ when constructing the latent mask $M$ (Sec.~\ref{subsec:latent-masking}).

Users often prefer backgrounds that are simple yet reflective of the document’s theme, without specifying a prompt for every page~\cite{weng2024desigen}. By extracting $s_i$ automatically, the system synthesizes backgrounds even when no user prompt $p$ is provided. This enables prompt-free, content-driven generation that is both minimal and document-specific~\cite{cui2021fbc}.

\subsection{Instruction Generation Model}
Given the semantic summary $s_i$ and/or a user prompt $p$, together with the context of prior pages $H_{i-1} = \{u_1, \dots, u_{i-1}\}$, we generate a page-level instruction $u_i$:
\begin{equation}
u_i = f_{\text{inst}}(s_i, p, H_{i-1}),
\end{equation}
where $f_{\text{inst}}$ produces a concise instruction describing a visual motif. Conditioning on $H_{i-1}$ allows—but does not strictly enforce—multi-page coherence by making style cues available to subsequent pages. Unlike video or dialogue, document backgrounds must remain visually coherent across pages while reflecting local content. Stateless prompting fails here, since independently generated instructions diverge stylistically. To mitigate this, we adapt the idea of a \emph{Recursive Narrative Bank}~\cite{kang2025action2dialogue} for documents, where a structured memory of prior page-level instructions conditions subsequent generations. We define the recursive memory at step $i$ as:
\vspace*{-1em}
\begin{equation}
H_i = \{ u_{i}^{(1)}, u_{i}^{(2)}, \dots, u_{i}^{(N)} \},
\end{equation}
where $u_{i}^{(k)}$ denotes prior instructions within a memory window of size $N$. The next instruction is generated as
\vspace*{-1em}
\begin{equation}
u_i = f_{\text{inst}}\big(s_i, p, H_{i-1}\big).
\end{equation}
This recursive structure allows accumulated stylistic cues (color tones, textures) to persist across the document. Visual coherence here denotes consistent reuse of global stylistic elements (palette, tone, motifs), while still allowing local variation tied to $s_i$.

\subsection{Foreground-Aware Latent Masking for Document Generation}
\label{subsec:latent-masking}
We first define the vanilla diffusion update. Let $x_t$ denote the latent state at time $t$ and $v_t^{\mathrm{raw}}$ the model-predicted velocity:
\begin{equation}
x_{t-\Delta t} \;=\; x_t \;-\; v_t^{\mathrm{raw}} \,\Delta t.
\end{equation}

\paragraph{Mask Construction.}
The latent is arranged on a 2D lattice with height $h$ and width $w$. We define
\begin{equation}
M_{ij} \;=\;
\begin{cases}
\lambda, & (i,j)\in \mathcal{C}(\rho;h,w),\\[4pt]
1, & \text{otherwise},
\end{cases}
\end{equation}
where $\mathcal{C}(\rho;h,w)$ is a centered window covering a fraction $\rho$ of the lattice; $\lambda\in(0,1)$ is the attenuation factor. Mapping $M$ onto token positions yields a mask $\mathbf{m}$.

\paragraph{Time-Gated Modulation.}
Masking is applied at later timesteps. The effective velocity becomes
\begin{equation}
v_t' = \mathbf{m}\odot v_t^{\mathrm{raw}} + (1-\mathbf{m})\odot \mathrm{stopgrad}(v_t^{\mathrm{raw}}),
\label{Eqn:TGM}
\end{equation}
and the update is
\begin{equation}
x_{t-\Delta t} = x_t - v_t' \,\Delta t.
\label{Eqn:TGM_update}
\end{equation}
This softly attenuates generation in text regions while keeping background areas rich and variable. 

\paragraph{Discussion.}
Foreground boxes (from layout analysis) provide precise regions for ARO, while representative layout boxes define the latent mask window. Related works on diffusion-based inpainting also apply region masks, but our time-gated, attenuation-based variant using Eqn.~\ref{Eqn:TGM} and Eqn.~\ref{Eqn:TGM_update} is tailored for document readability. Unlike amplification-based strategies, we deliberately suppress foreground updates, stabilizing text regions while letting the background evolve.

\subsection{Automated Readability Optimization (ARO)}
\label{subsec:aro}
Instead of adding uniform opaque backing shapes to ensure readability, ARO computes the minimal opacity $\alpha^*$ of semi-transparent backings that meets WCAG~2.2~\cite{wcag22_w3c2025} contrast.

\paragraph{Contrast Calculation.}
Convert sRGB to linear RGB:
\begin{equation}
C_{\mathrm{lin}} =
\begin{cases}
C_{\mathrm{srgb}}/12.92, & C_{\mathrm{srgb}}\le 0.04045,\\[6pt]
\big((C_{\mathrm{srgb}}+0.055)/1.055\big)^{2.4}, & \text{otherwise}.
\end{cases}
\end{equation}
Relative luminance:
\vspace*{-1em}
\begin{equation}
L(R,G,B) = 0.2126R_{\mathrm{lin}} + 0.7152G_{\mathrm{lin}} + 0.0722B_{\mathrm{lin}}.
\end{equation}
WCAG contrast:
\vspace*{-1em}
\begin{equation}
CR(L_1,L_2) = \frac{\max(L_1,L_2)+0.05}{\min(L_1,L_2)+0.05}.
\end{equation}

\paragraph{Opacity Search.}
For overlay luminance $L_o$ and background pixel $L_{\mathrm{bg}}$, blended luminance:
\begin{equation}
L_{\mathrm{blend}}(\alpha) = \alpha L_o + (1-\alpha) L_{\mathrm{bg}}.
\end{equation}
Minimal opacity $\alpha^*$ is
\vspace*{-1em}
\begin{equation}
\alpha^* = \min \left\{ \alpha \,\bigg|\,
\frac{1}{N}\sum_{i=1}^N \mathbf{1}\!\left[ CR\!\big(L_{\mathrm{blend}}^{(i)}(\alpha), L_t\big) \geq \tau \right] \ge \rho \right\}.
\end{equation}

Final opacity:
\begin{equation}
\alpha = \min\big(1,\,\max(\alpha^*+\epsilon,\,\alpha_{\min})\big).
\end{equation}

\paragraph{Overlay Construction.}
Each bounding box $(\mathbf{b}_{i,\ell}) \in \mathcal{L}_i$ is expanded and drawn as a rounded rectangle with adaptive overlay color and computed opacity. This RGBA overlay is composited above the background. 

\paragraph{Discussion.}
ARO ensures readability with minimal intervention. Unlike prior opaque patches, it adapts to local luminance, producing natural, semi-transparent backings.

\subsection{Document generation and interactivity}

\begin{figure*}[t]
    \centering
    \includegraphics[width=\linewidth]{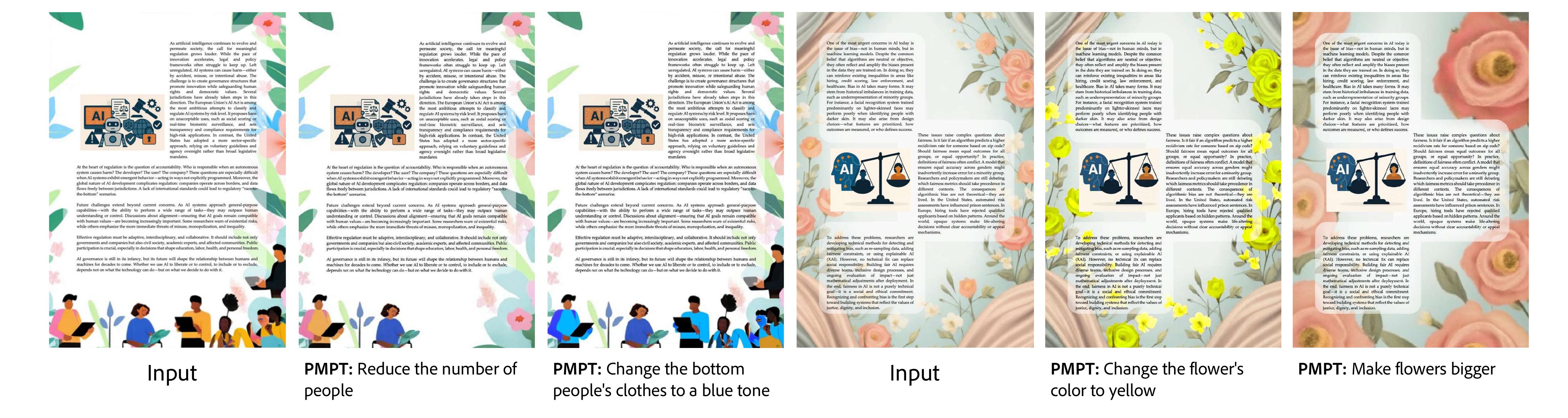}
    \vspace*{-1.5em}
    \caption{\textbf{Feedback-based document editing.} Our system enables post-generation refinement through prompts. Users can modify only the background layer—without altering text or figures— such as reducing the number of people, adjusting colors, style and scale. }
    \label{fig:doc_editing}
    \vspace*{-2em}
\end{figure*}

As shown in Fig.~\ref{fig:doc_editing}, our system supports both automatic and interactive use. In automatic mode, $\{s_i\}$ and $\{u_i\}$ drive background generation with masking+ARO. In interactive refinements (e.g., “make the background more subtle”), we recompute $u_i$ and regenerate without modifying text. For structural edits (text content changes), layout boxes are recomputed and the full pipeline reruns with updated masking and ARO. 

\section{Experiments and Results}
\subsection{Benchmarking Document Datasets}
\label{sec:dataset}
To evaluate our framework under controlled yet realistic conditions, we constructed a benchmark consisting of academic-style documents and slide decks. Each instance contains exactly three pages or slides, designed to simulate instructional materials with coherent narrative progression. This structure enables assessment of both per-page visual quality and cross-page stylistic consistency, which is central to background editing in multi-page settings. Each page integrates long-form text, structured bullet points, and at least one embedded figure positioned in non-trivial layouts (e.g., side-anchored, floating, or interleaved with text). These layouts introduce realistic spatial constraints for foreground preservation while avoiding artificial simplifications. By fixing foreground content and layout, the benchmark directly reflects the constrained editing scenario studied in this work. All textual and visual content is authored or generated specifically for this benchmark (text via GPT-5 and images via GPT-4o) to ensure full reproducibility and redistribution. \textbf{Constructing a publicly shareable benchmark from real academic papers, textbooks, or presentation slides has limitations, as such materials typically cannot be freely modified or redistributed.} Publicly available “real-world” document corpora are often domain-specific (e.g., government notices or news articles), which would introduce unintended dataset bias. Our synthetic construction therefore enables balanced coverage across layout styles, thematic topics, and page-level narrative structures while maintaining legal clarity and experimental control. Importantly, the benchmark focuses on document editing scenarios where background synthesis is meaningful—namely, materials with intentional margins, structured layouts, and moderate text density typical of lecture slides and academic handouts. We intentionally avoid severely degraded scanned documents or OCR-heavy artifacts, as such settings shift the problem toward document restoration rather than foreground-preserving background generation. This design isolates the core challenges addressed in this paper—layout fidelity, readability preservation, and multi-page stylistic coherence—without conflating them with unrelated noise factors. Additional implementation details and dataset statistics are provided in Appendix~\ref{appendix:dataset}.

\vspace*{-1em}
\subsection{Implementation Details}
For both PDFs (A4) and slides (16:9), empty dummy images were used as inputs to ensure format alignment, upon which the Summarization Model distilled document text into concise visual themes and the Instruction Generation Model generated editing prompts under multi-page consistency, both model implemented with GPT-4o; bounding boxes were extracted with PyMuPDF and OpenCV, where precise text-region boxes were supplied to ARO for contrast-aware overlays (target contrast 7.0, coverage 0.98, padding 24, radius fraction 0.12), while representative bounding boxes that enclose the overall text regions were provided to Latent Masking to guide attenuation (strength=0.2, start step=0.29 of the diffusion schedule). Unlike ARO, which requires pixel-accurate boxes to guarantee WCAG contrast, Latent Masking only needs representative boxes since its goal is to suppress intrusive background synthesis rather than ensure pixel-level legibility. Latent Masking was implemented inside BAGEL, which serves as a baseline framework for background editing, and our contributions build upon it without altering the novelty of our readability-preserving and thematically consistent generation pipeline. We note that GPT-4o is used only in the Summarization and Instruction modules, which treat the LLM as a swappable component; LM, ARO, and RNB are LLM-independent. The \emph{Ours w/o MPC} ablation isolates this dependency: removing the LLM-driven module changes only consistency (CLIP MP $-0.06$, LLM Voting $-0.07$) while WCAG ($99.69\%$) and OCR ($0.96$) remain unchanged (Tab.~\ref{tab:quantitative_results}).

\vspace*{-1em}
\subsection{Qualitative Results} 
\vspace*{-0.75em}
We evaluate our framework against two representative baselines: BAGEL~\cite{deng2025emerging}, a document-oriented editing system, and GPT-5~\cite{openai2025introducinggpt5}, a strong general-purpose multimodal model. Our task differs fundamentally from layout synthesis or poster generation approaches~\cite{chen2025posta,zhang2025creatiposter}, which typically assume flexible structural rearrangement or text re-rendering. In contrast, we address a constrained editing setting in which the \textbf{foreground text, figures, and layout of an existing multi-page document must remain unchanged, and only the background layer is modified.} We therefore \textbf{restrict comparisons to methods that can operate directly on fixed document pages without regenerating layout elements or retypesetting text.} Including layout-generation pipelines would require architectural adaptation and alter the problem definition, making direct comparison misleading. Figures~\ref{fig:results_pdfs_representative}--\ref{fig:results_slides_representative} demonstrate that our method preserves structural fidelity and readability while introducing visually coherent backgrounds across styles and page sequences. BAGEL often produces strong textures or high-frequency patterns that partially interfere with dense text regions. GPT-5 can generate aesthetically appealing outputs, but frequently alters existing layout structure, modifies embedded figures, or hallucinates additional text, which violates the foreground-preservation constraint central to our task. Across both Prompt+Text and Prompt-only modes, our approach maintains layout integrity while achieving stylistic consistency across pages. Additional qualitative examples and further discussion on baseline selection are provided in Appendix~\ref{appendix:qualitative}.

\begin{table*}[t]
\vspace*{-1em}
\centering
\small
\resizebox{\textwidth}{!}{%
\begin{tabular}{lccccccccc}
    \toprule
    \textbf{Method} 
    & \textbf{Layout} $\uparrow$ 
    & \textbf{Color} $\uparrow$ 
    & \textbf{Graphic Style} $\uparrow$ 
    & \textbf{Compliance} $\uparrow$ 
    & \textbf{WCAG} $\uparrow$ (\%) 
    & \textbf{OCR Acc.} $\uparrow$ 
    & \textbf{CLIP MP Consistency} $\uparrow$ 
    & \textbf{CLIP Prompt Score} $\uparrow$ 
    & \textbf{LLM Voting} $\uparrow$ \\
    \midrule
    BAGEL & 4.1025 & 4.275 & 4.1671 & 4.325 & 66.98 & 0.5536 & 0.5571 & 0.1877 & 4.2292 \\
    GPT-5 & 3.8807 & 4.0685 & 4.0050 & 4.1164 & 55.02 & 0.5225 & 0.6870 & 0.1687 & 4.0185 \\
    \textbf{Ours (Prompt+Text mode)} & 4.2028 & 4.4285 & 4.2485 & 4.4000 & \textbf{99.75} & \textbf{0.9665} & \textbf{0.6955} & \textbf{0.2374} & \textbf{4.3342} \\
    \textbf{Ours (Prompt only mode)} & \textbf{4.355} & \textbf{4.545} & \textbf{4.3478} & \textbf{4.7357} & 99.38 & 0.9578 & 0.6259 & 0.2042 & 4.5100 \\
    \midrule
    Ours w/o LM  & 4.2700 & 4.5407 & 4.3528 & 4.7100 & 99.67 & 0.9085 & 0.6905 & \textbf{0.2542} & 4.4907 \\
    Ours w/o ARO & 4.0835 & 4.4114 & 4.2057 & 4.4992 & 97.35 & 0.9012 & 0.6886 & 0.2336 & 4.3214 \\
    Ours w/o MPC & 4.1664 & 4.3807 & 4.2107 & 4.1985 & 99.69 & 0.9632 & 0.6420 & 0.2287 & 4.2592 \\
    \bottomrule
\end{tabular}}
\caption{\textbf{Quantitative evaluation of document background generation across nine metrics.} LLM-judged metrics (Layout, Color, Graphic Style, Compliance, LLM Voting) are evaluated on a 1--5 scale by GPT-4o, while automatic metrics (WCAG Contrast Coverage, OCR Accuracy, CLIP MP Consistency, CLIP Prompt Score) measure text readability, accessibility compliance, multi-page consistency, and image--text alignment. Results are reported for BAGEL, GPT-5, and our method in two operating modes (Prompt+Text and Prompt-only), as well as internal ablations. Higher values indicate better performance for all metrics.}
\label{tab:quantitative_results}
\vspace*{-2em}
\end{table*}

\begin{figure}[hb]
\vspace*{-1.6em}
    \centering
    \includegraphics[width=0.7\linewidth]{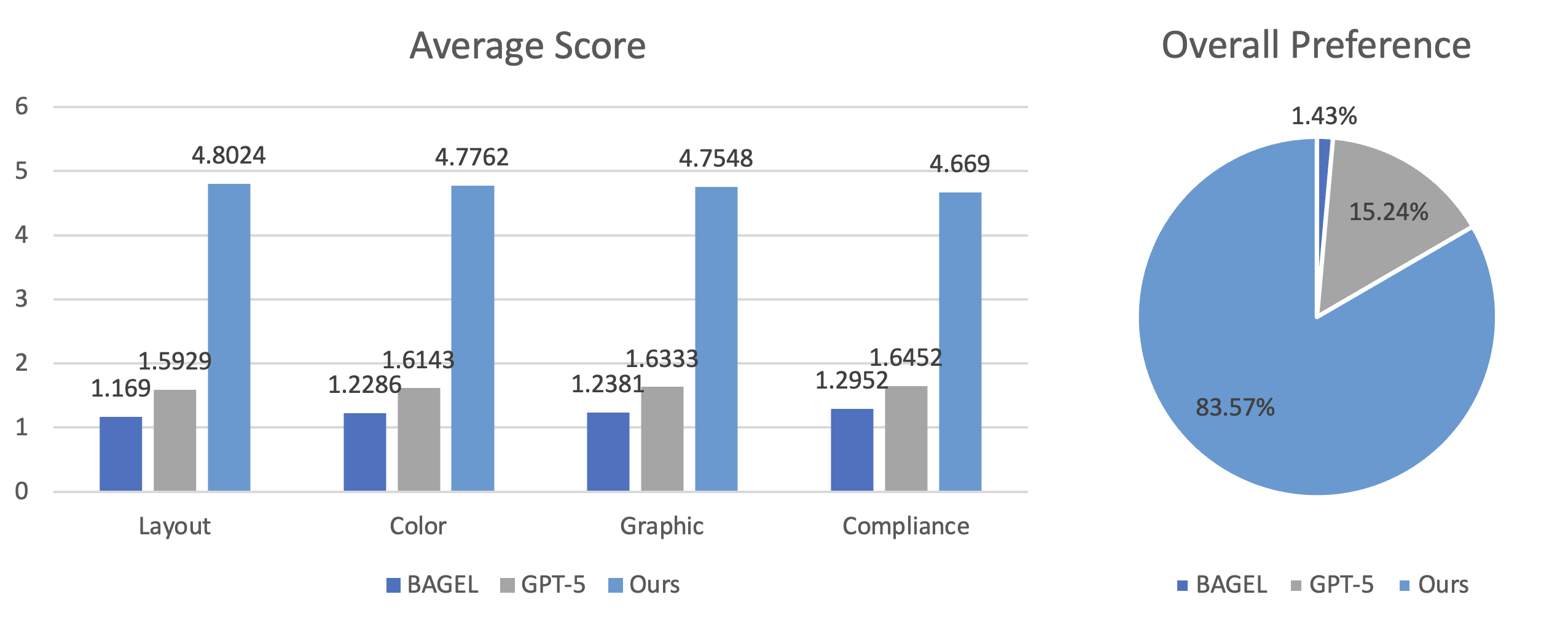}
    \vspace*{-1em}
    \caption{
    \textbf{User study results.} Thirty participants evaluated three anonymized systems across four design dimensions: Layout, Color, Graphic Style, and Prompt Compliance (left). Our method achieved the highest score in all categories. In overall preference voting (right), \textbf{83.57\%} of users selected our result over GPT-5 (15.24\%) and BAGEL (1.43\%).}
    \vspace*{-1em}
    \label{fig:user_study}
\end{figure}

\subsection{Quantitative Evaluation}
We evaluate against BAGEL and GPT-5 using metrics aligned with prior document-generation work~\cite{zhang2025creatiposter}. Our method achieves the best performance across all categories: (i) Design Quality (layout, color, graphic style, compliance),  (ii) Readability (WCAG contrast coverage, OCR accuracy),  (iii) Multi-page Consistency (CLIP similarity, LLM voting). Our framework reaches near-perfect WCAG compliance (99.75\%) and the highest CLIP-based consistency.  Detailed analysis are shown in Appendix~\ref{appendix:quantitative}.

\subsection{User Study}
We conducted a user study with 30 participants to evaluate our document background generation framework. For each task, participants viewed the original document (PDF or slide) and three anonymized outputs (BAGEL, GPT-5, and \emph{Ours}) and rated them across four design dimensions—\emph{Layout preservation}, \emph{Color harmony}, \emph{Graphic style consistency}, and \emph{Prompt compliance}—using a 5-point Likert scale. Appropriate measures were taken to ensure that method identities were not disclosed to participants during the study. As shown in Figure~\ref{fig:user_study}, our method achieved the highest score in every dimension (4.669–4.8024), whereas BAGEL and GPT-5 scored substantially lower (1.169–1.6452). Participants also selected their preferred output for each task, and \textbf{83.57\%} of all votes favored \emph{Ours}, compared to 15.24\% for GPT-5 and 1.43\% for BAGEL. These results demonstrate that users consistently prefer our approach for maintaining readability and producing visually coherent backgrounds.  Detailed analysis and explanations are provided in Appendix~\ref{appendix:user_study}.

\vspace*{-1em}
\subsection{Ablation Study}
\vspace*{-0.5em}
We validate the contributions of each component: latent masking (LM), automated readability optimization (ARO), and multi-page consistency (MPC). Removing LM or ARO reduces readability (WCAG and OCR). Removing MPC reduces thematic continuity across pages while preserving readability. Detailed analysis are included in Appendix~\ref{appendix:ablation}.

\vspace*{-1em}
\section{Conclusion}
\vspace*{-1em}
In this work, we introduced a multi-layered document editing framework with automated text-conditioned background design to generate visually coherent, text-preserving, and thematically consistent multi-page documents. Our method applies latent masking, which protects text and figures through soft attenuation in the diffusion space, inspired by smooth barrier functions in physics and numerical optimization. This strategy suppresses updates in sensitive regions without applying hard erasure, allowing backgrounds to be regenerated while preserving readability. To further guarantee legibility, we integrate Automated Readability Optimization (ARO), a contrast-driven, WCAG-guided algorithm that adaptively determines the minimal opacity of semi-transparent backings per text region. ARO ensures that inserted shapes remain visually harmonious with the background while meeting accessibility standards, and in combination with latent masking, yields both natural and readable results. Furthermore, we ensure multi-page visual consistency through a recursive summarization-and-instruction process, where each page is distilled into a compact representation that guides subsequent generations, enabling coherent evolution of visual motifs across entire slide decks or reports. Finally, by adopting a layered editing paradigm that treats text, figures, and backgrounds as separate compositional elements, and incorporating prompt-based customization, the framework balances automated coherence with flexible user control, bridging generative modeling and real-world document design workflows. \textbf{Limitations and Future Works.} Detailed discussion of limitations and future work is provided in Appendix~\ref{app:limitations}.

\noindent

\bibliographystyle{splncs04}
\bibliography{main}

\newpage
\onecolumn
\appendix
\section{Appendix}

\subsection{Ethics Statement}
\label{supplementary:ethics}

\begin{tcolorbox}[breakable, title=Ethics Statement]
All documents and slides (textual content and images) used in our evaluation were generated using GPT-4o's multimodal image generation capabilities and GPT-5's text generation capabilities. No external images, web-scraped materials, or third-party resources were used.

Our dataset does not contain any personal, sensitive, or identifiable information about real individuals, nor does it reference confidential or private documents. Although no human data appears in the dataset itself, we conducted a separate user study to evaluate usability and preference of generated document backgrounds. This study received \textbf{IRB exemption under the category of minimal-risk research} at our institution, confirming that the procedures comply with human subjects research ethics standards. The content topics (e.g., history, science education, ethics) were intentionally chosen to avoid harmful or sensitive subject matter. For document and background generation, we ensured that the system does not fabricate misleading factual claims or intentionally alter document meaning during editing. The proposed model performs background synthesis \emph{without modifying, obscuring, or removing existing text or foreground content}. All experiments respect the principle of preserving authorial intent and maintaining readability.

We acknowledge that background manipulation systems may potentially be misused—for example, to conceal information or modify documents deceptively. To mitigate this concern, our design strictly restricts operations to background regions, prevents edits to textual content, and encourages transparent versioning by requiring explicit user instructions for edits. Overall, the dataset and experiments adhere to responsible AI research principles: respect for intellectual property, transparency of provenance, and prevention of harm through misuse.
\end{tcolorbox}

\subsection{Limitations and Future Works}
\label{app:limitations}
While we introduce several novel contributions, some limitations remain. 

First, latent masking, while effective for preserving textual fidelity, is not always perfect. Since it attenuates updates in the diffusion process instead of completely prohibiting them, residual artifacts can appear around text boundaries, especially in dense, irregular layouts, and ARO reliability degrades on irregular transparent figure overlays. Foreground extraction also relies on PyMuPDF/OpenCV, so dense mathematical content or nested tables can cause missed regions; these yield graceful readability degradation rather than foreground corruption. This trade-off reflects the design choice of suppressing content generation rather than erasing it outright, and although it preserves readability in most cases, errors may emerge. Second, the summarization-and-instruction mechanism occasionally oversimplifies nuanced content, which can limit semantic alignment between visual motifs and document meaning. Third, our benchmark fixes documents at three pages to isolate cross-page consistency under tractable human evaluation; although the Recursive Narrative Bank is length-agnostic by construction (per-page $O(1)$ memory), its fixed window saturates beyond roughly ten pages, and empirical validation on long-form documents remains future work. Fourth, we do not report an empirical SD-Inpainting baseline, since a fair configuration at document resolutions (A4, 16:9) is non-trivial---downsampling destroys the text legibility being evaluated, while tiling introduces implementation-dependent seams. 

Conceptually, inpainting synthesizes new content into binary mask regions, whereas our latent masking softly attenuates updates so the background evolves around protected regions; the \emph{Ours w/o LM} ablation ($\text{OCR }0.91$/$\text{WCAG }99.67\%$) already bounds such inpainting-style behavior from above. Finally, while user prompts provide stylistic flexibility, finer-grained control, such as per-section themes or adaptive palette shifts, remains an open challenge. 

Future work could explore more adaptive masking strategies, enhanced semantic encoders for tighter content--visual alignment, scaling to long-form documents, and interactive editing pipelines where users refine backgrounds iteratively across multi-page documents; we further discuss commercial deployment scenarios and their requirements in Appendix~\ref{app:commercial}.

\subsection{Commercial Use Cases and Real-World Impact}
\label{app:commercial}
Our contribution is not the discovery of a new use case but a technical enabler for already-deployed ones. Commercial tools such as Canva Magic Design and Microsoft Designer expose automated background generation and may benefit from algorithms that ensure foreground readability when applied to text-dense documents such as reports, slides, and handouts. Our framework directly targets this scenario: as shown in Table~\ref{tab:quantitative_results}, it reaches OCR $0.97$ / WCAG $99.75\%$, compared with $0.55$/$66.98\%$ (BAGEL) and $0.52$/$55.02\%$ (GPT-5)---a near-saturation readability regime that our baselines do not approach. The $83.57\%$ user-preference margin (Fig.~\ref{fig:user_study}; $N{=}30$, a mixed pool of researchers, designers, and graduate students) further indicates that this is the operating point users prefer when our outputs are presented alongside strong commercial-grade alternatives. As future work, integrating readability-preserving background generation into such production design pipelines---where foreground text fidelity and accessibility compliance are first-class constraints---represents a natural and high-impact direction.

\subsection{Experiment Input Prompts}
\label{app:prompts}

To evaluate the robustness of our framework across diverse stylistic intentions, we prepared representative user prompts covering seven categories: \emph{geometric}, \emph{shapes}, \emph{textures}, \emph{colorful}, \emph{muted}, \emph{professional}, and \emph{real and natural objects}. Table~\ref{tab:input_prompts} lists the prompts used in our experiments. Although the proposed system conceptually supports \textbf{optional user guidance} — allowing backgrounds to be generated solely from the page summary ($s_i$) — in our experiments we evaluate two practical modes:

\begin{itemize}
    \item \textbf{Prompt + Text mode}: The Instruction Model receives both the page summary $s_i$ and the user prompt $p$, and produces a page-level instruction that balances document grounding and user-specified style.
    \item \textbf{Prompt only mode}: The Instruction Model receives only $p$, without access to $s_i$, representing style-driven generation without content grounding.
\end{itemize}

Both settings allow user prompts to be minimal and high-level (e.g., ``Add a cream background with snowflakes''), rather than requiring detailed designs. While the system architecture enables a fully prompt-free setting (automatic instruction generation from $s_i$), we focus our evaluation on the two modes above, which represent realistic usage patterns observed in document editing scenarios.

\begin{table*}[t]
\centering
\begin{tabular}{|l|p{0.75\linewidth}|}
\hline
\textbf{Category} & \textbf{Example user prompt} \\
\hline
Geometric & ``Add a background a modern abstract background of layered geometric forms, built from clean, repeating patterns in harmonious alignment with precise symmetry for a sleek visual effect'' \\
\hline
Shapes & ``Add a background a playful yet balanced arrangement of varied shapes, blending bold curves and soft angles into a dynamic composition with natural depth'' \\
\hline
Textures & ``Add a background a richly detailed background where contrasting textures, from smooth to coarse, layer together to create tactile depth and engaging visual interest'' \\
\hline
Colorful & ``Add a background a lifelike scene filled with a diverse range of vivid hues, each rendered under natural lighting to interact dynamically and create a vibrant atmosphere'' \\
\hline
Muted & ``Add a background a softly lit, realistic setting with a gentle, desaturated palette where subdued colors evoke calmness and timeless elegance'' \\
\hline
Professional & ``Add a background a refined, realistic design with minimal clutter, clean lines, and understated details, balanced by lighting for a polished, professional tone'' \\
\hline
Real and natural objects & ``Add a background a bright and inviting lifelike setting inspired by real-world elements, incorporating subtle everyday details for an authentic, harmonious feel'' \\
\hline
\end{tabular}
\caption{Experiment input prompts across seven style categories. Prompts are optional: when omitted, the Summarization LLM provides content-driven grounding; when supplied, prompts act as soft stylistic constraints for the Instruction Model.}
\label{tab:input_prompts}
\end{table*}

\subsection{Detailed Benchmarking Document Datasets}
\label{appendix:dataset}
To rigorously evaluate our framework, we constructed a collection of demonstration datasets in two formats: academic documents and academic-style slides. Each instance consists of exactly three pages, designed to emulate realistic educational and scholarly materials. In the case of slides, each set was structured according to a conventional presentation flow of introduction, body, and conclusion, while documents adopt a similarly coherent three-part thematic organization. This design ensures that generated backgrounds are tested not only on isolated pages, but also on sequences where topical progression and narrative continuity must be preserved.  

All materials were carefully curated to balance text-heavy content with bullet-point highlights, reflecting the layout of typical university lecture slides or reports. This dual use of dense paragraphs and concise key points captures the range of textual styles commonly encountered in academic practice, offering an effective benchmark for readability-preserving generation. Each page additionally contains at least one image, placed in non-trivial arrangements to increase layout complexity and provide realistic stress tests for spatial preservation.  

All textual and image content was generated by GPT-4o, with multimodal image generation capabilities, to avoid using third-party copyrighted content, while capturing the stylistic qualities of real instructional materials. By combining free-form but academically inspired layouts with thematic progression across three-page units, the dataset offers a controlled yet realistic testbed for evaluating text-preserving and thematically consistent background generation. For our qualitative evaluation, we employed these datasets in both PDF (A4 format) and slide (16:9 PPT format) settings.

\paragraph{Corpus overview.}
We propose two parallel benchmarking corpora designed to stress-test background generation under realistic, complicated layouts:
(1) \emph{Academic Documents} (7 topics; PDF), and
(2) \emph{Academic Slides} (7 topics; 16:9 slide decks).
Every file contains exactly three pages/slides. Each page includes \emph{dense paragraph text (documents)} or \emph{bullet points (slides)}, plus \emph{at least one image} deliberately placed to create non-trivial text–figure interactions. Slides follow a strict \textbf{Introduction $\rightarrow$ Body $\rightarrow$ Conclusion} structure; documents follow a coherent three-part thematic organization with heavier prose, mirroring university handouts/readings.

\paragraph{Copyright compliance and provenance.}
All textual content and images were generated with ChatGPT-4o (text and multimodal image generation). No other external or third-party content is used.

\paragraph{Design goals.}
Unlike template-constrained benchmarks, our materials are \emph{free-form by design}. We intentionally vary:
(i) text density (long paragraphs vs.\ concise bullets),
(ii) visual placement (left/right columns, inset figures, wrap-around text),
(iii) semantic focus across pages (P1 vs.\ P2 vs.\ P3),
to evaluate whether background generation remains \emph{text-preserving}, \emph{layout-aware}, and \emph{theme-consistent} in cluttered or irregular settings. The mixture of prose and bullets aids both human study replicability and automated scoring (readability, OCR, contrast).

\subsubsection{Academic Documents (7 topics; 3 pages each)}
\begin{itemize}
  \item \textbf{(D1) The Legacy of Ancient Civilizations} \\
  P1. Mesopotamia and the Invention of Writing, P2. Egyptian Art and Afterlife, P3. Greek Influence on Modern Politics
  \item \textbf{(D2) Understanding Volcanoes} \\
  P1. Volcanic Eruption Mechanisms, P2. Famous Historical Eruptions, P3. Monitoring and Risk Management
  \item \textbf{(D3) Women in Modern Science} \\
  P1. Scientific Breakthroughs by Women, P2. Barriers and Gender Disparities, P3. Policy \& Programs Supporting Women
  \item \textbf{(D4) The Cultural Impact of the Olympic Games} \\
  P1. The Olympics as a Global Stage, P2. Politics and Protest in the Olympics, P3. Commercialization and Media Spectacle
  \item \textbf{(D5) Visit South Korea: Beyond K-POP} \\
  P1. Historical Sites and UNESCO Heritage, P2. Regional Food and Culinary Culture, P3. Contemporary Arts and Global Presence
  \item \textbf{(D6) Introduction to AI Ethics} \\
  P1. Philosophical Foundations, P2. Algorithmic Bias and Fairness, P3. Regulation and Future Challenges
  \item \textbf{(D7) Understanding Sleep} \\
  P1. Biological Functions of Sleep, P2. Sleep Disorders and Modern Lifestyles, P3. Improving Sleep Quality
\end{itemize}

\subsubsection{Academic Slides (7 decks; 16:9; 3 slides each)}
Each deck uses the same narrative scaffold to induce page-level semantic shifts for background conditioning:
\emph{Slide 1 = Introduction}, \emph{Slide 2 = Body (analysis/mechanisms/cases)}, \emph{Slide 3 = Conclusion (implications/strategy/outlook)}.
All slides combine paragraph text \emph{and} bullet points, plus one or more figures.

\begin{itemize}
\item \textbf{(S1) Climate Change and Global Policy} \
P1. Scientific Basis of Climate Change, P2. International Agreements and Treaties, P3. Adaptation and Mitigation Strategies

\item \textbf{(S2) The Ethics of Gene Editing} \
P1. Introduction to Gene Editing Ethics, P2. Ethical Dilemmas and Case Studies, P3. Toward Responsible Innovation

\item \textbf{(S3) Renewable Energy Transition} \
P1. The Urgency of Transition, P2. Challenges and Solutions, P3. A Sustainable Future

\item \textbf{(S4) Space Exploration and Humanity’s Future} \
P1. Why We Explore Space, P2. Current Missions and Technologies, P3. The Ethical and Strategic Horizon

\item \textbf{(S5) The Psychology of Decision-Making} \
P1. Introduction to Decision Science, P2. Cognitive Biases in Action, P3. Improving Decision Quality

\item \textbf{(S6) Cybersecurity in a Connected World} \
P1. The Growing Importance of Cybersecurity, P2. Threats and Vulnerabilities, P3. Strategies for Resilience

\item \textbf{(S7) Climate Change and Public Health} \
P1. Climate Change as a Health Crisis, P2. Health Impacts and Inequalities, P3. Integrating Climate and Health Policy
\end{itemize}

\paragraph{Layout diversity.}
Across both corpora, we vary: single- vs.\ dual-column layouts for documents; image placement (left/right rail on pages); and text structure (paragraphs in documents vs.\ bullet lists in slides). Images on slides always appear below their captions, whereas document figures may be adjacent to surrounding text. This diversity forces background models to (i) preserve text legibility, (ii) avoid overpainting foreground figures/logos, and (iii) maintain page-specific thematic focus.

\paragraph{Intended use.}
The datasets serve as a controlled yet realistic testbed for:
\emph{(a)} foreground preservation (text, figures, margins),
\emph{(b)} background–content alignment per page/slide (changing key points across P1–P3),
\emph{(c)} multi-page stylistic coherence without template overfitting,
and \emph{(d)} automated readability/contrast/OCR evaluation.

\subsection{Detailed Qualitative Results}
\label{appendix:qualitative}
We present qualitative comparisons of our method against two state-of-the-art document editing models: BAGEL~\cite{deng2025emerging} and GPT-5~\cite{openai2025introducinggpt5}. A key distinction of our problem setting is that the \emph{input is an existing document page (PDFs or slides)}, where text regions and layout must be preserved. Our task is thus defined as \emph{document-centric background generation}: given an existing page, we synthesize visually coherent backgrounds that improve readability and ensure multi-page consistency, while strictly maintaining the foreground content. This setting is inherently an \emph{editing} problem rather than generation from scratch. Therefore, our setting differs fundamentally from poster generation frameworks such as POSTA~\cite{chen2025posta} and CreatiPoster~\cite{zhang2025creatiposter}, which take only text as input from the user rather than existing documents, and from general-purpose text-to-image models, which generate new images, not refer to existing documents, from text prompts and thus fall outside the scope of document-centric editing. In selecting baselines, we required models that (1) accept arbitrary aspect ratios and resolutions as an input to match document page sizes, (2) enable explicit background \emph{editing} instead of full-scene synthesis, and (3) are publicly accessible. Under these criteria, BAGEL and GPT-5 remain the only models suitable for direct comparison.

\begin{figure*}[t]
    \centering
    \includegraphics[width=\linewidth]
    {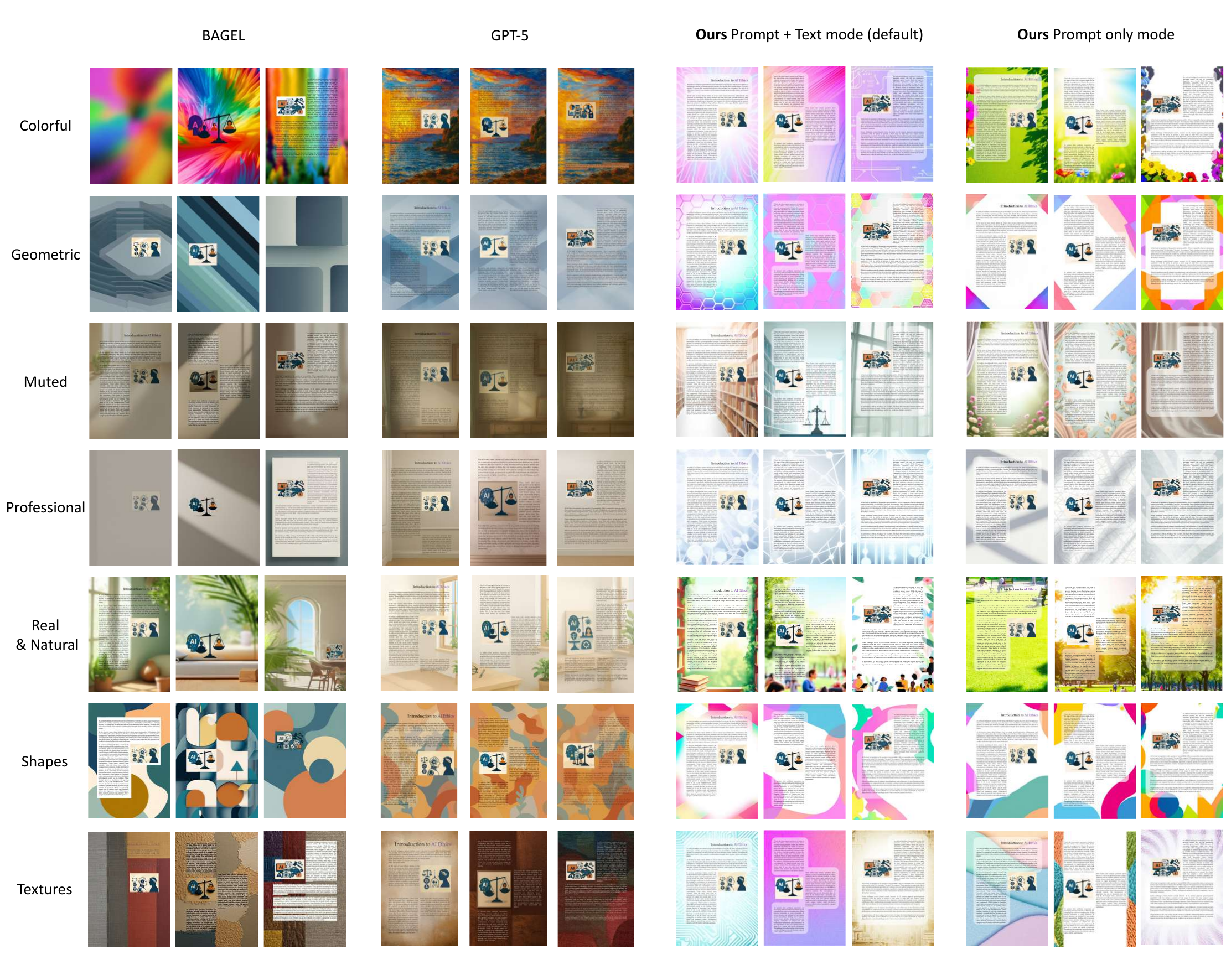}
    \caption{Qualitative comparison on academic-style \textbf{PDFs} (A4). Rows correspond to style conditions (Colorful, Geometric, Muted, Professional, Real \& Natural, Shapes, Textures).}
    \label{fig:results_pdfs}
\end{figure*}

\begin{figure*}[t]
    \centering
    \includegraphics[width=\linewidth]{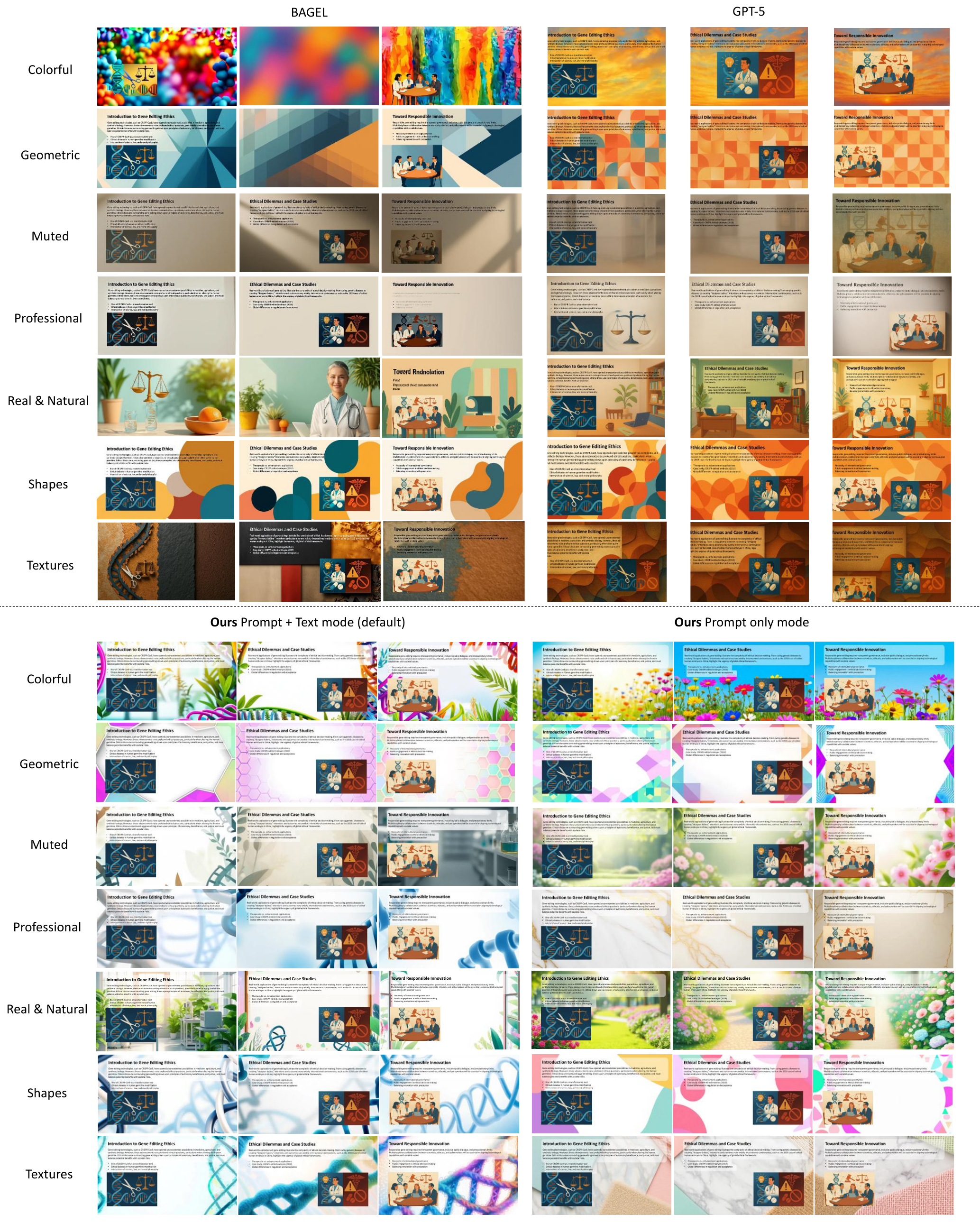}
    \caption{Qualitative comparison on academic-style \textbf{slides} (16:9). Rows correspond to style conditions (Colorful, Geometric, Muted, Professional, Real \& Natural, Shapes, Textures). }
    \label{fig:results_slides}
\end{figure*}

\begin{figure*}[t]
    \centering
    \includegraphics[width=\linewidth]{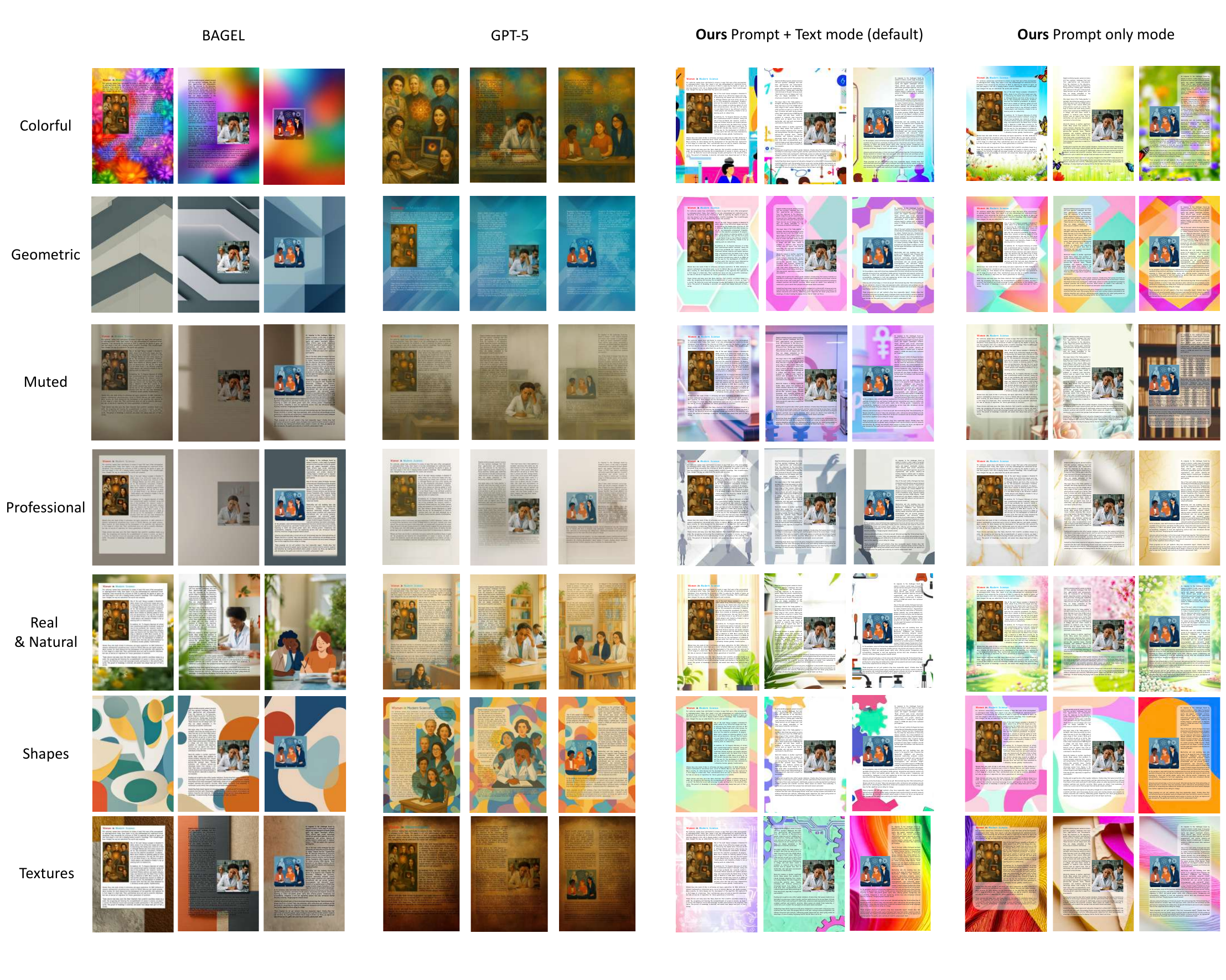}
    \caption{Additional qualitative comparison on academic-style \textbf{PDFs} (A4). Rows correspond to style conditions (Colorful, Geometric, Muted, Professional, Real \& Natural, Shapes, Textures).}
    \label{fig:results_pdfs_supp}
\end{figure*}

\begin{figure*}[t]
    \centering
    \includegraphics[width=\linewidth]{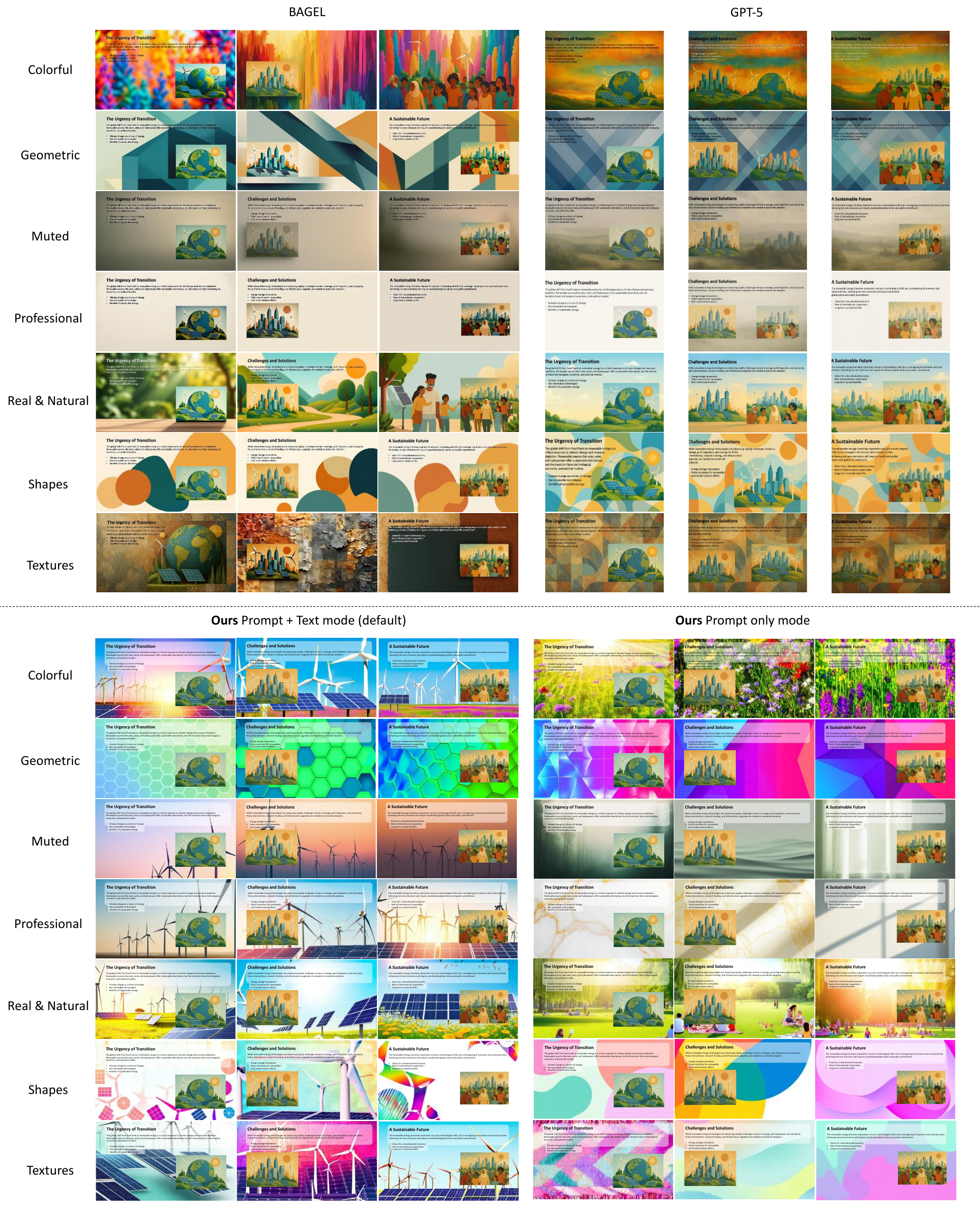}
    \caption{Additional qualitative comparison on academic-style \textbf{slides} (16:9). Rows correspond to style conditions (Colorful, Geometric, Muted, Professional, Real \& Natural, Shapes, Textures). }
    \label{fig:results_slides_supp}
\end{figure*}

Figures~\ref{fig:results_pdfs}, \ref{fig:results_slides}, \ref{fig:results_pdfs_supp} and \ref{fig:results_slides_supp} present qualitative results on academic-style PDFs and slides under multiple stylistic conditions (Colorful, Geometric, Muted, Professional, Real \& Natural, Shapes, Textures). Both variants of our model—with and without the Summarization Model—consistently maintain readability, layout fidelity, and multi-page visual consistency across diverse formats. 

We focus our evaluation on two practically relevant usage modes commonly observed in document editing workflows—(1) users provide a stylistic prompt, and (2) users rely solely on the document content. These settings reflect realistic interaction patterns while covering the full capability space of our framework. For our default setting, Prompt + Text mode (w/ Summarization Model), generated backgrounds are more explicitly aligned with the dominant themes of each page, yielding semantically grounded and context-aware motifs. For Prompt only mode (w/o Summarization Model), the system still preserves consistency across pages through the Instruction Generation pipeline, producing visually coherent outputs that remain non-intrusive to text regions. In both cases, the framework harmonizes stylistic variation with content protection, demonstrating robustness across document types. 

These results indicate that our approach reliably balances thematic alignment, stylistic coherence, and readability preservation, regardless of whether summarization is applied. We compared our framework against baseline methods such as BAGEL, highlighting improvements in naturalness, consistency, and readability.

\paragraph{On the scope of baseline selection.}
Recent advances in graphic design and layout-aware generation~\cite{peng2025bizgen, zhang2025creatidesign, wu2025hybrid} primarily address content synthesis from structured specifications, such as layout annotations, bounding boxes, or textual design briefs. In contrast, our problem formulation assumes a fully instantiated multi-page document in which foreground text, figures, and layout are immutable. The objective is not to synthesize or rearrange structural elements, but to enhance the background layer while preserving existing content and accessibility constraints. As a result, direct comparison with layout-generation systems would require re-defining the task to allow structural modification or re-rendering of text, thereby departing from the constrained editing scenario we study. For a controlled and meaningful evaluation, we therefore limit baselines to methods that can operate on arbitrary document pages as input, preserve layout fidelity, and support background-level modification without regenerating foreground content. Under these requirements, BAGEL and GPT-5 constitute representative and practically deployable baselines for document-centric background editing. For completeness, we also considered recent strong image-generation models such as Nano Banana~2. While capable of high-quality explanatory figure generation, it does not provide a layout-preserving editing interface: given a document page, it tends to reinterpret the input as a photographed document and distort foreground text rather than modifying only the background layer. This is the same exclusion criterion that applies to GPT-4o/5 and layout-generation pipelines---the limitation is one of task definition rather than model capability.

\subsection{Detailed Quantitative Analysis}
\label{appendix:quantitative}
We evaluate our framework against two baselines, BAGEL and GPT-5, using eight quantitative metrics grouped into three categories: design quality, readability, and multi-page consistency. All metrics are reported at the document level, with higher scores indicating better performance unless otherwise specified. 

\textbf{Design Quality.} We adopt four dimensions of evaluation that have been widely recognized in prior work on poster and document generation \cite{zhang2025creatiposter}: layout, color, graphic style, and compliance. Following this methodology, an LLM (GPT-5) is prompted to assign scores on a 1--5 scale for each dimension. Layout measures the balance and appropriateness of text and element placement, Color assesses harmony between palettes and content, and Graphic Style captures the stylistic consistency of fonts, decorative elements, and backgrounds. Compliance, which in CreatiPoster was judged solely by LLM, is extended in our evaluation by also reporting a separate CLIP-based prompt similarity score as an additional quantitative reference. In practice, compliance is still judged by GPT-4o to reflect nuanced prompt adherence, while the CLIP score serves as a complementary, embedding-level check. Our method consistently achieves the highest scores across all four dimensions, with Layout (4.2028), Color (4.4285), Graphic Style (4.2485), and Compliance (4.40), outperforming both BAGEL and GPT-5. 

\textbf{Readability.} We assess document accessibility through two complementary measures. WCAG Contrast Coverage computes the percentage of text regions whose luminance contrast ratio meets the WCAG~2.2 AA threshold of 4.5:1, ensuring that backgrounds do not hinder legibility. OCR Accuracy is measured at the character level using Tesseract by aligning OCR outputs with embedded text references in the PDF. Our framework achieves 99.75\% WCAG compliance and 0.97 OCR accuracy, substantially higher than BAGEL (66.98\%, 0.55) and GPT-5 (55.02\%, 0.52). These results confirm the combined effectiveness of latent masking, which prevents background intrusion into foreground regions, and Automated Readability Optimization (ARO), which adaptively adjusts opacity to achieve perceptual contrast while preserving aesthetics. 

\textbf{Multi-page Consistency.} Thematic coherence across consecutive pages is measured using CLIP-based Consistency and LLM Voting. The CLIP-based measure computes cosine similarity between background embeddings of adjacent pages, while LLM Voting assigns a score from 1--5 based on overall continuity of motifs and styles across a document. Our framework achieves the strongest results, with a CLIP consistency of 0.70 and an LLM voting score of 4.33, outperforming BAGEL (0.56, 4.23) and GPT-5 (0.69, 4.02). The relatively small gap in CLIP scores compared to GPT-5 contrasts with the larger improvement in LLM voting, suggesting that our recursive summarization and instruction mechanism better preserves high-level thematic consistency beyond local visual similarity. We note that our framework supports two user-selectable operating modes—\emph{Prompt+Text} (with summarization) and \emph{Prompt-only}. Since this toggle reflects user intent (whether backgrounds should reflect document text or remain purely stylistic), we report both for completeness but do not treat them as ablations; all ablations in this paper vary only internal modules (LM, ARO, MPC). Unless otherwise stated, the default operating mode is \emph{Prompt+Text}.

To ensure fair and tractable evaluation, we follow the dataset selection strategy adopted in prior document generation studies such as POSTA \cite{chen2025posta}. Specifically, we reserve one genre from both the PDF and slide sets for validation, corresponding to approximately one-sixth of the data in each category. This split maintains a representative evaluation scale without requiring the entire dataset, balancing coverage and efficiency. Overall, our framework (LM + ARO + MPC) achieves the best performance across all eight metrics. The largest improvements are observed in readability, where near-perfect WCAG compliance and OCR accuracy are attained, and in multi-page consistency, where recursive context propagation yields coherent motifs across entire documents. These results demonstrate the effectiveness of adapting diffusion to document-centric background generation while preserving accessibility and consistency (Table~\ref{tab:quantitative_results}).

\subsection{Detailed Ablation Study}
\label{appendix:ablation}
To disentangle the contributions of each module in our framework, we conduct ablation experiments under the default Prompt+Text mode. Specifically, we remove (i) latent masking (LM), (ii) automated readability optimization (ARO), and (iii) multi-page consistency (MPC), while keeping all other components fixed. Results are summarized in Table~\ref{tab:quantitative_results}. Removing latent masking results in a sharp decline in readability metrics. WCAG compliance drops from 99.75\% to 99.67\% and OCR accuracy falls from 0.97 to 0.91. This confirms that LM plays a critical role in preventing background textures from spilling into text regions. Interestingly, CLIP Prompt Score increases slightly (0.25 vs.\ 0.24), suggesting that stronger background updates can sometimes enhance text–image alignment, but at the cost of text legibility. Without ARO, WCAG compliance falls more substantially to 97.35\%, and OCR accuracy drops to 0.90. This indicates that ARO’s contrast-aware opacity adjustment is crucial for meeting accessibility standards. Design quality scores remain largely comparable, but the visual harmony of text backings degrades, explaining the small decreases in Color and Graphic Style ratings. When multi-page consistency is disabled, readability remains strong (WCAG 99.69\%, OCR 0.96), but cross-page coherence degrades. CLIP MP Consistency falls from 0.70 to 0.64, and LLM Voting drops from 4.33 to 4.26. These results highlight MPC’s role in propagating motifs across pages, improving high-level coherence without compromising text preservation. Each component improves distinct aspects of document background generation: LM and ARO jointly ensure readability and accessibility, while MPC enhances thematic consistency. Together, these modules yield the strongest overall performance, demonstrating that our framework is most effective when all three are combined.

\subsection{Detailed User Study}
\label{appendix:user_study}
To assess the effectiveness of our document background generation framework, we conducted a human subject study with 30 participants. Each participant was presented with the original document (either a PDF page or a slide) together with three anonymized outputs corresponding to BAGEL, GPT-5, and \emph{Ours}. The method identities were concealed throughout the study to prevent brand or familiarity bias. For each task, participants evaluated the three outputs across four graphic design dimensions: \emph{Layout preservation}, \emph{Color harmony}, \emph{Graphic style consistency}, and \emph{Prompt compliance}. Each dimension was rated on a 5-point Likert scale (1 = strongly disagree, 5 = strongly agree). All participants completed evaluations on 14 document tasks covering diverse layout and style conditions. The presentation format and evaluation instructions were kept consistent across participants to ensure comparability. Although the display order of the three anonymized outputs was fixed for all participants, method names were not revealed, and participants were explicitly instructed to evaluate each output independently based on visual quality and readability criteria. Figure~\ref{fig:user_study} reports aggregated scores across all participants and tasks. Our method achieved the highest rating in every evaluation dimension, with mean scores of 4.8024 (Layout), 4.7762 (Color), 4.7548 (Graphic Style), and 4.669 (Prompt Compliance), while BAGEL and GPT-5 scored considerably lower (BAGEL: 1.169–1.2952, GPT-5: 1.5929–1.6452 on average). Participants additionally selected an overall preferred result for each task. Across the 14 tasks, 83.57\% of preferences were assigned to \emph{Ours}, compared with 15.24\% for GPT-5 and 1.43\% for BAGEL. These results indicate a strong user preference for outputs that preserve document structure while achieving stylistic coherence.

\paragraph{Survey Setup and Interface.}
Participants were presented with 14 evaluation tasks (7 PDF pages and 7 slides).  For each task, the original document was shown along with three generated background–edited versions, labeled as Document A, Document B, and Document C. Each result corresponded to one of the three systems (BAGEL, GPT-5, and \emph{Ours}), but the model identities were not disclosed to participants. To help participants understand the intended visual concept of the document without requiring them to read the full text, two additional pieces of context were provided alongside the input document:

\begin{itemize}
    \item \textbf{Input Prompt} – the user prompt used for background generation (e.g., ``Colorful'', ``Muted'', ``Professional'').
    \item \textbf{Content Summary} – a short semantic summary automatically generated by our Summarization Model, describing the main topics of the page (e.g., ``AI fairness and bias concerns'').
\end{itemize}

These were shown so that participants could evaluate how well each generated background matched the intended visual style and document content, without needing to inspect long text passages. No free-form comments were collected; each evaluation focused solely on the four quantitative dimensions.

\paragraph{Conditions.}
The 14 evaluation tasks covered seven background style categories used in the paper: \textit{Colorful}, \textit{Geometric}, \textit{Muted}, \textit{Professional}, \textit{Real \& Natural}, \textit{Shapes}, and \textit{Textures}. Each style was evaluated once for a PDF page and once for a slide.

\paragraph{Protocol and IRB Compliance.}
The study took approximately 15--20 minutes per participant. No personal information was collected, and participants were free to discontinue at any time.  This study was reviewed by the Institutional Review Board (IRB) at the University of Maryland (IRB\#2368547-1) and determined to be IRB Exempt, as no personally identifiable or sensitive information was collected, and all evaluation data were anonymized.

\paragraph{Quantitative Summary.}
Across all dimensions (\emph{Layout, Color, Graphic Style, Prompt Compliance}), our method achieved the highest ratings, with mean scores ranging from \textbf{4.669–4.8024}. BAGEL and GPT-5 scored significantly lower (averaging \textbf{1.169–1.6452}). In the per-task overall preference selection, \textbf{83.57\%} of participants’ votes favored \emph{Ours}, compared with 15.24\% for GPT-5 and 1.43\% for BAGEL. These results demonstrate that users strongly preferred our method for preserving readability and producing visually coherent, style-consistent backgrounds.

\subsection{Computation Time and Memory Consumption}
\label{supp:compute}

\begin{table}[htb!]
    \small
    \centering
    \resizebox{0.95\textwidth}{!}{
    \begin{tabular}{l|c|c}
    \toprule
    \textbf{Configuration (3-page document)} 
    & \textbf{Peak GPU Memory (GB)} 
    & \textbf{Inference Time (sec)} \\
    \midrule
    BAGEL
    & 31.3461 GB / 40.000 GB 
    & 246.00 sec \\
    \midrule
    GPT-5 (API-based) 
    & N/A 
    & 324.00 sec \\
    \midrule
    Ours (LM only) 
    & 31.3461 GB / 40.000 GB 
    & 246.00 sec \\
    \midrule
    Ours (LM + ARO) 
    & 31.3461 GB / 40.000 GB 
    & 247.26 sec \\
    \midrule
    Ours (LM + ARO + MPC) 
    & 31.3461 GB / 40.000 GB 
    & \textbf{254.56 sec} \\
    \bottomrule
    \end{tabular}
    }
    \caption{\textbf{Computation time and memory consumption.}
    We report peak GPU memory usage and wall-clock inference time for generating backgrounds for a 3-page document on a single NVIDIA A100 GPU (40GB).
    All configurations use the same diffusion backbone.
    GPT-5 is accessed via a closed-source API; therefore, internal GPU memory consumption is not observable and reported as N/A.}
    \label{tab:compute-module}
\end{table}

Table~\ref{tab:compute-module} summarizes runtime and peak GPU memory measured on a single NVIDIA A100 (40GB) for full 3-page document generation.

\textbf{BAGEL}
The diffusion backbone requires 31.3461~GB peak GPU memory and 246~seconds for a 3-page document (82 seconds per page). As expected, diffusion sampling dominates total runtime due to repeated UNet forward passes across timesteps.

\textbf{GPT-5 (API-based baseline).}
GPT-5 is accessed through a hosted API service. End-to-end wall-clock latency (324 seconds) is measured at the client side. However, internal GPU memory usage is not exposed by the API and is therefore reported as N/A. Unlike our constrained background-editing formulation, GPT-5 performs holistic multimodal editing.

\textbf{Ours (LM only).}
Applying Latent Masking (LM) does not change peak GPU memory usage and does not increase runtime relative to the backbone (246 seconds). This is because LM performs in-place attenuation of latent updates without introducing additional forward passes.

\textbf{Ours (LM + ARO).}
Automated Readability Optimization (ARO) introduces 1.26 seconds of additional processing for a 3-page document. Since ARO operates purely in image space after diffusion sampling, GPU memory usage remains unchanged.

\textbf{Ours (LM + ARO + MPC).}
The full model requires 254.56 seconds for a 3-page document. The additional 7.30 seconds stem from LLM-based summarization and instruction generation used for multi-page consistency. This preprocessing does not modify diffusion memory usage and does not alter per-timestep complexity.

Overall, diffusion sampling remains the dominant computational cost (246 seconds).ARO contributes 1.26 seconds (0.5\% overhead), and MPC contributes 7.30 seconds (2.97\% overhead), resulting in a total increase of approximately 3.5\% relative to the backbone. Peak GPU memory usage remains identical across all configurations (31.3461 GB), indicating that the proposed components do not increase diffusion memory requirements.

\newpage

\begin{figure}[t]
    \centering
    \includegraphics[width=\linewidth]{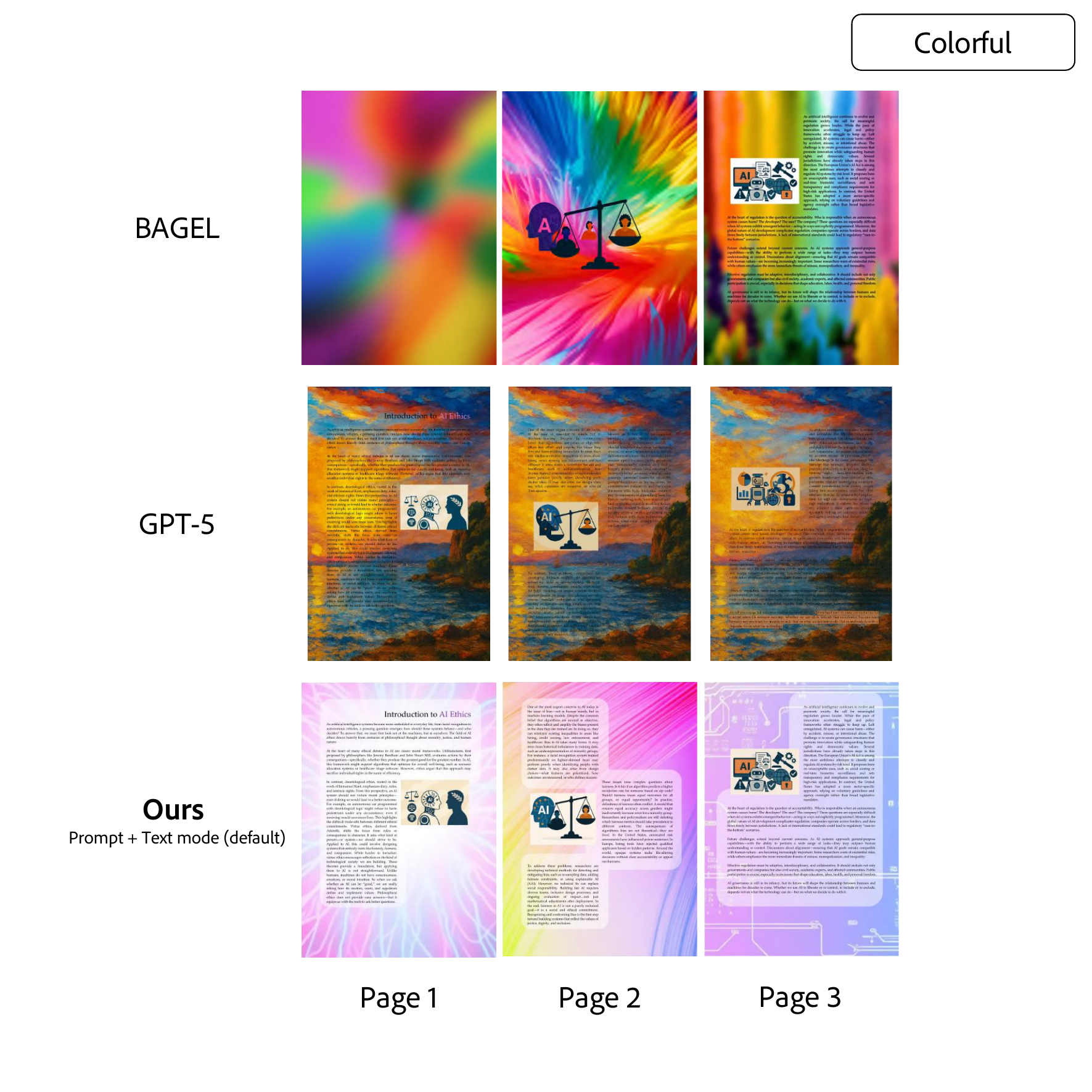}
    \caption{Comparison of background generation under the \emph{Colorful} style PDFs.}
\end{figure}

\newpage

\begin{figure}[t]
    \centering
    \includegraphics[width=\linewidth]{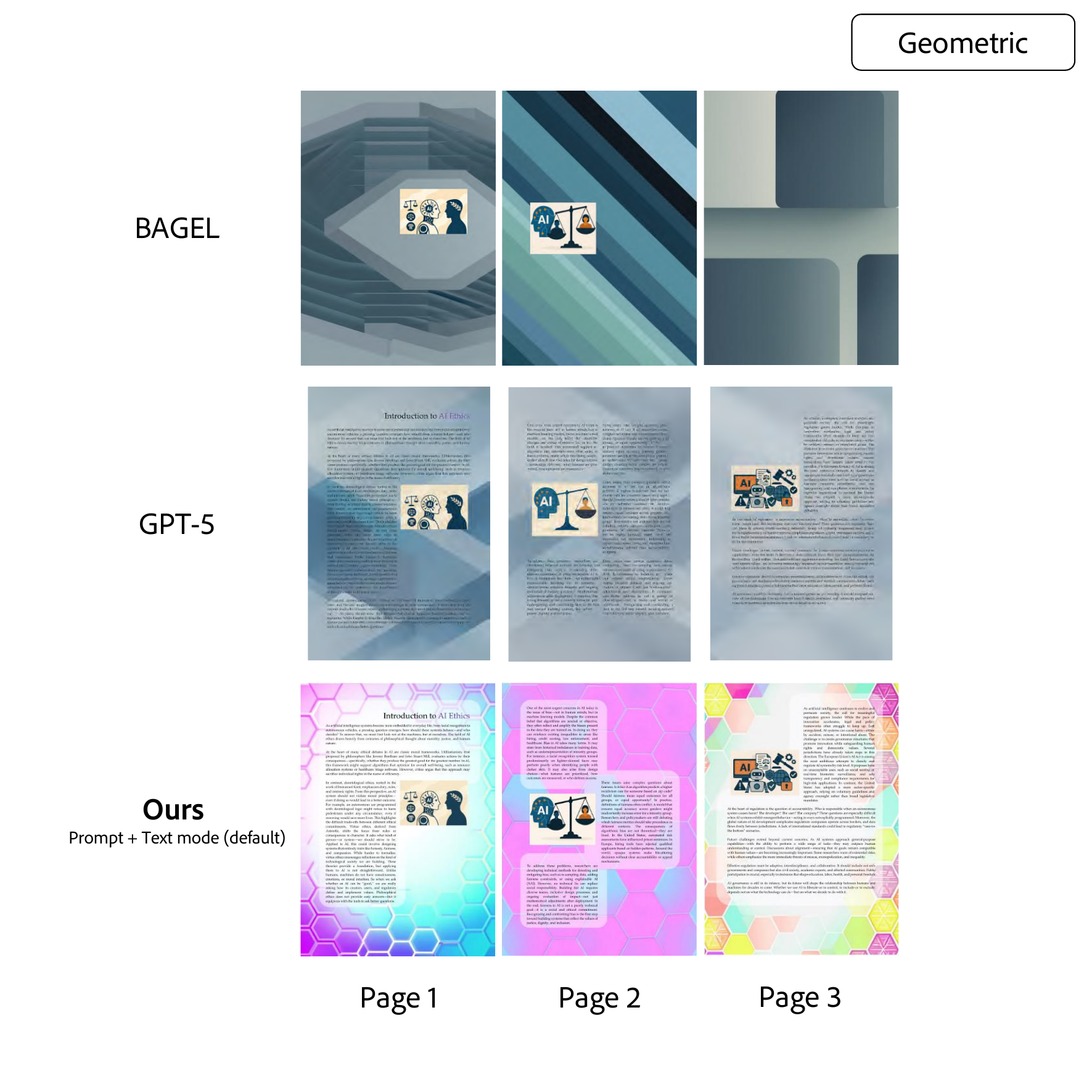}
    \caption{Comparison of background generation under the \emph{Geometric} style PDFs.}
\end{figure}

\newpage

\begin{figure}[t]
    \centering
    \includegraphics[width=\linewidth]{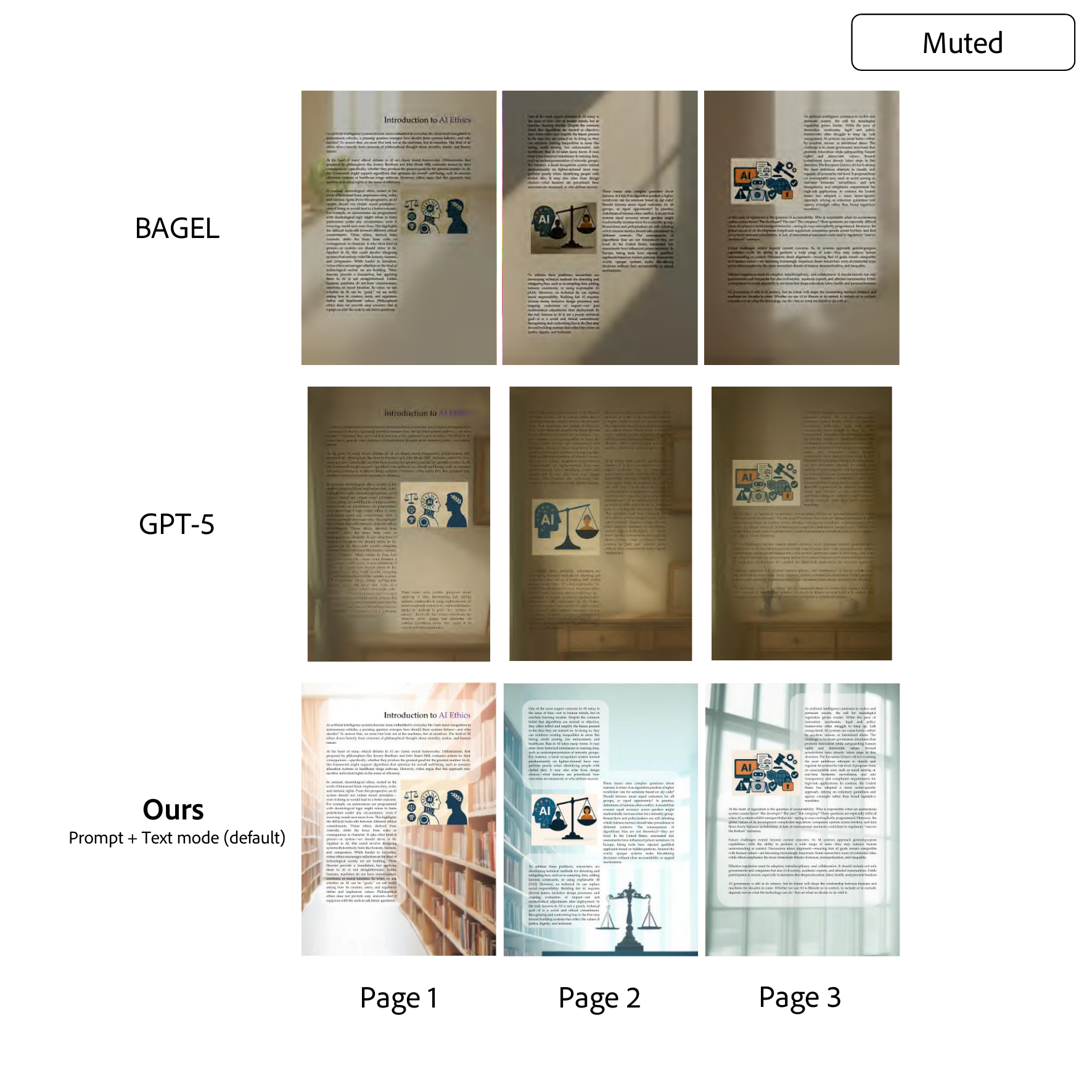}
    \caption{Comparison of background generation under the \emph{Muted} style PDFs.}
\end{figure}

\newpage

\begin{figure}[t]
    \centering
    \includegraphics[width=\linewidth]{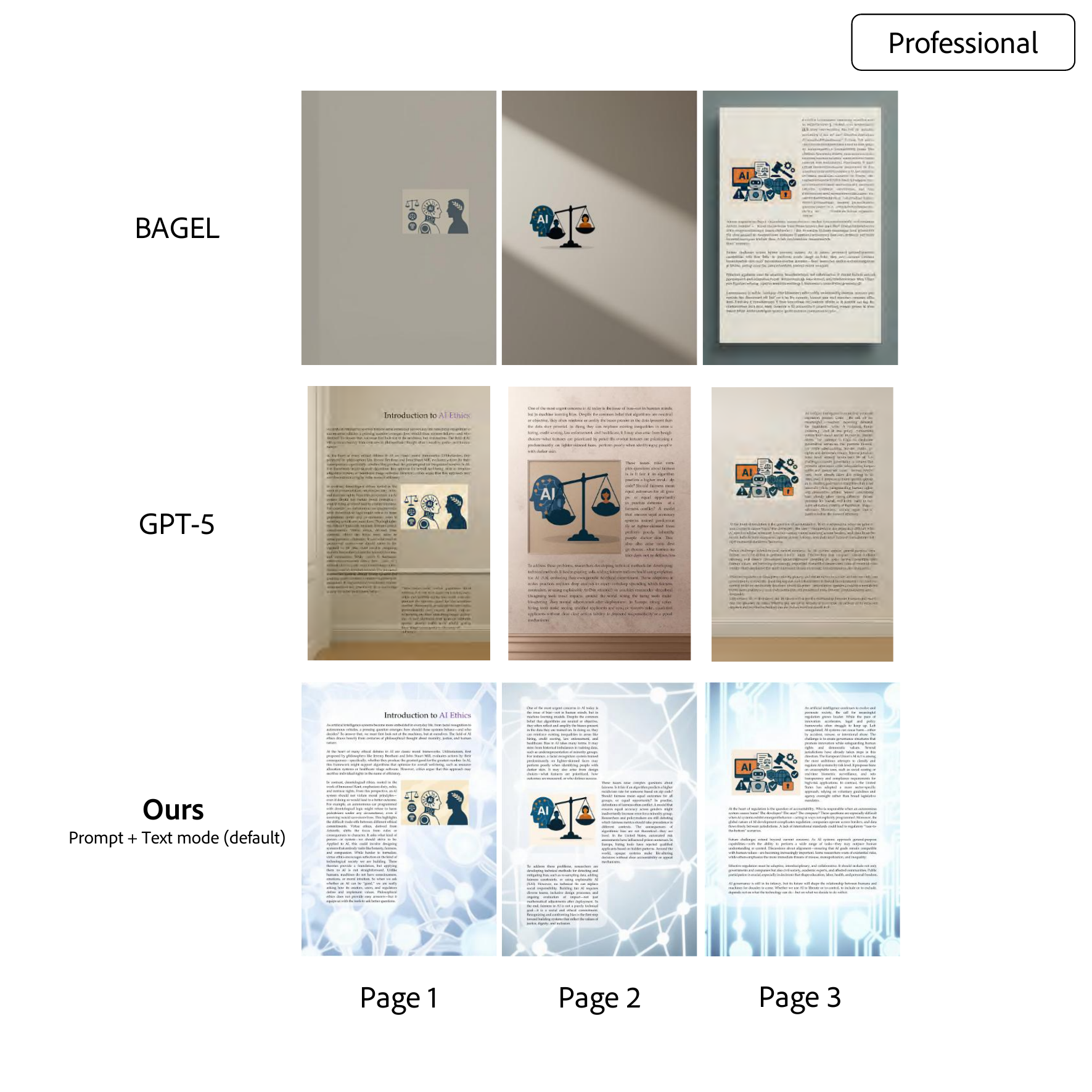}
    \caption{Comparison of background generation under the \emph{Professional} style PDFs.}
\end{figure}

\newpage

\begin{figure}[t]
    \centering
    \includegraphics[width=\linewidth]{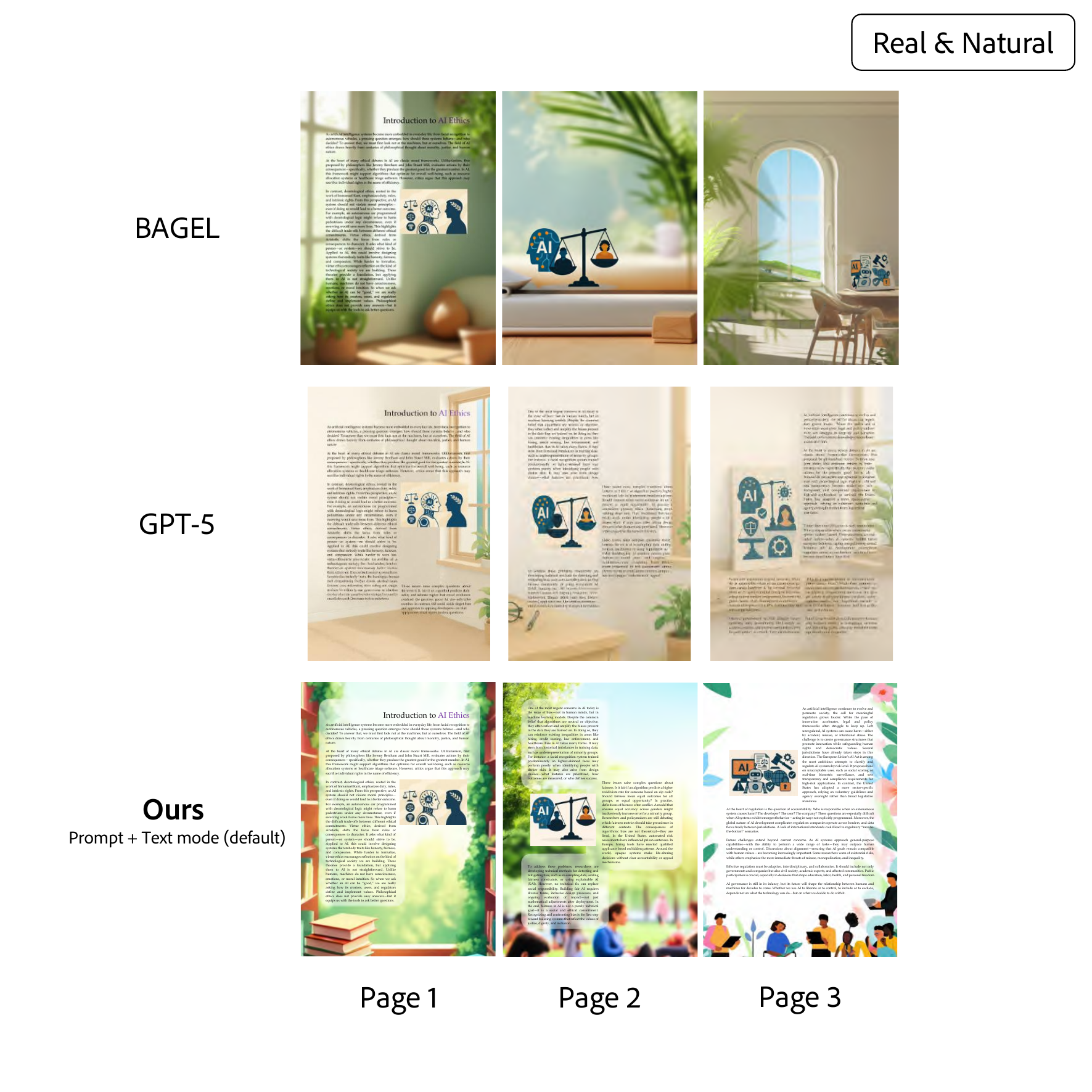}
    \caption{Comparison of background generation under the \emph{Real \& Natural} style PDFs.}
\end{figure}

\newpage

\begin{figure}[t]
    \centering
    \includegraphics[width=\linewidth]{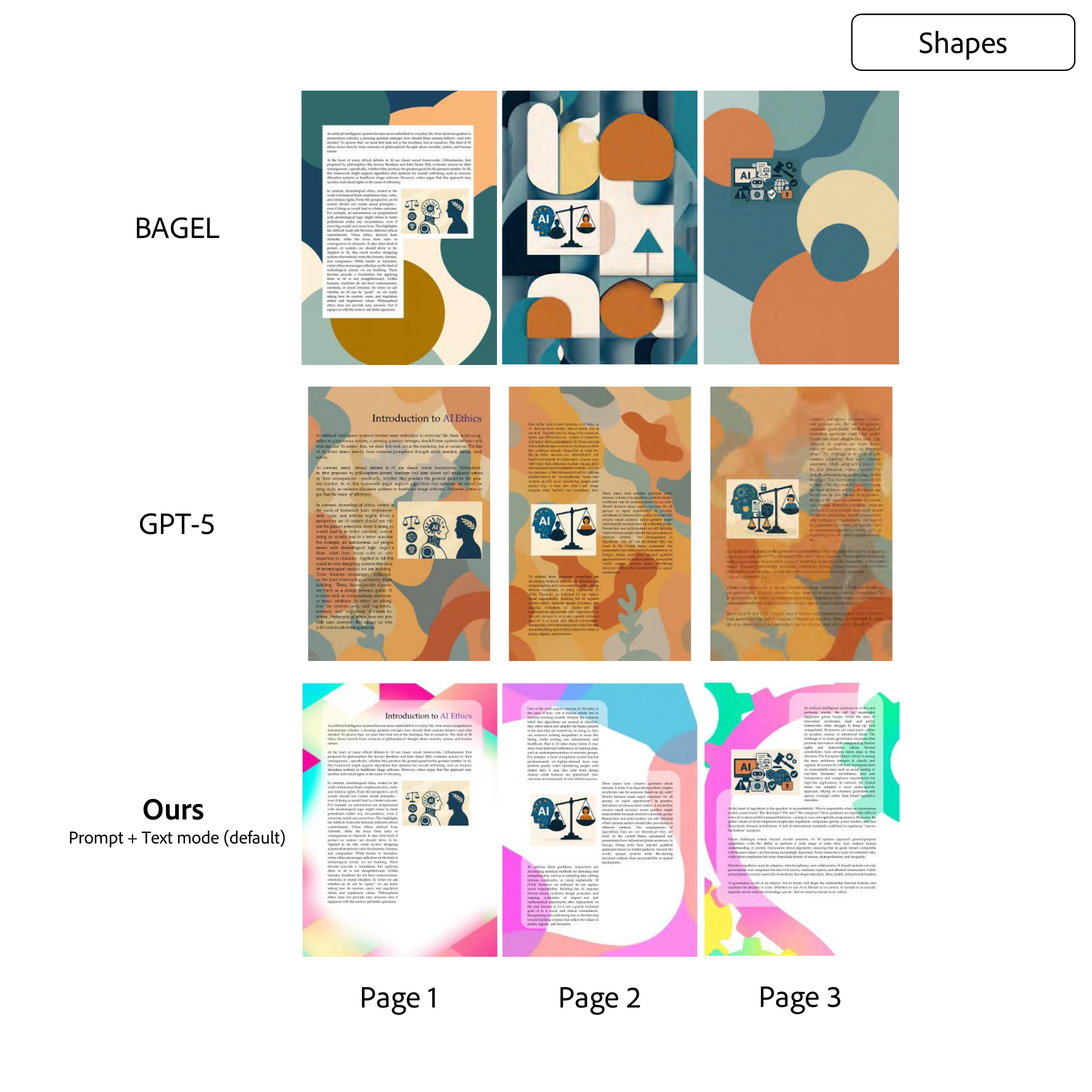}
    \caption{Comparison of background generation under the \emph{Shapes} style PDFs.}
\end{figure}

\newpage

\begin{figure}[t]
    \centering
    \includegraphics[width=\linewidth]{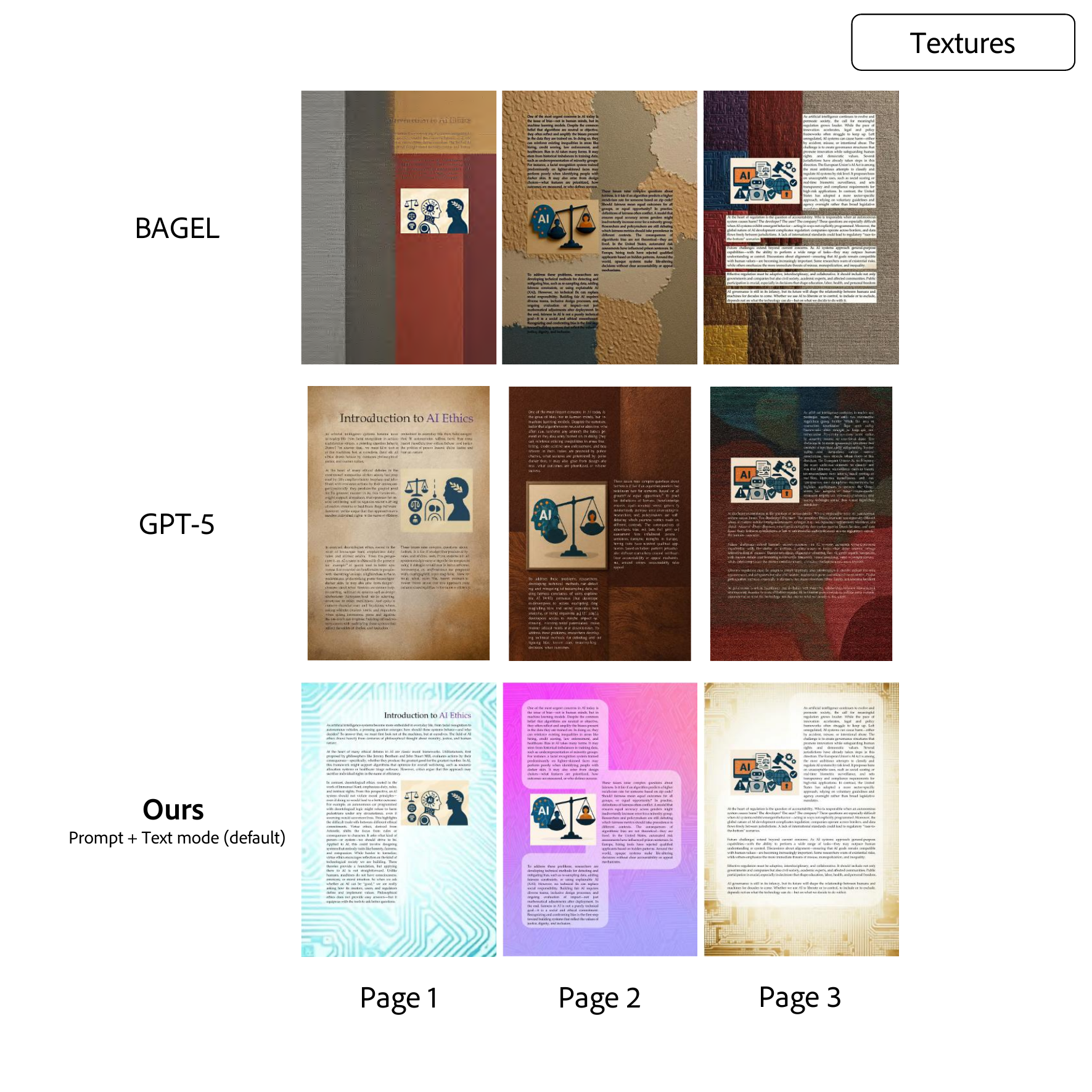}
    \caption{Comparison of background generation under the \emph{Textures} style PDFs.}
\end{figure}

\newpage

\begin{figure}[t]
    \centering
    \includegraphics[width=\linewidth]{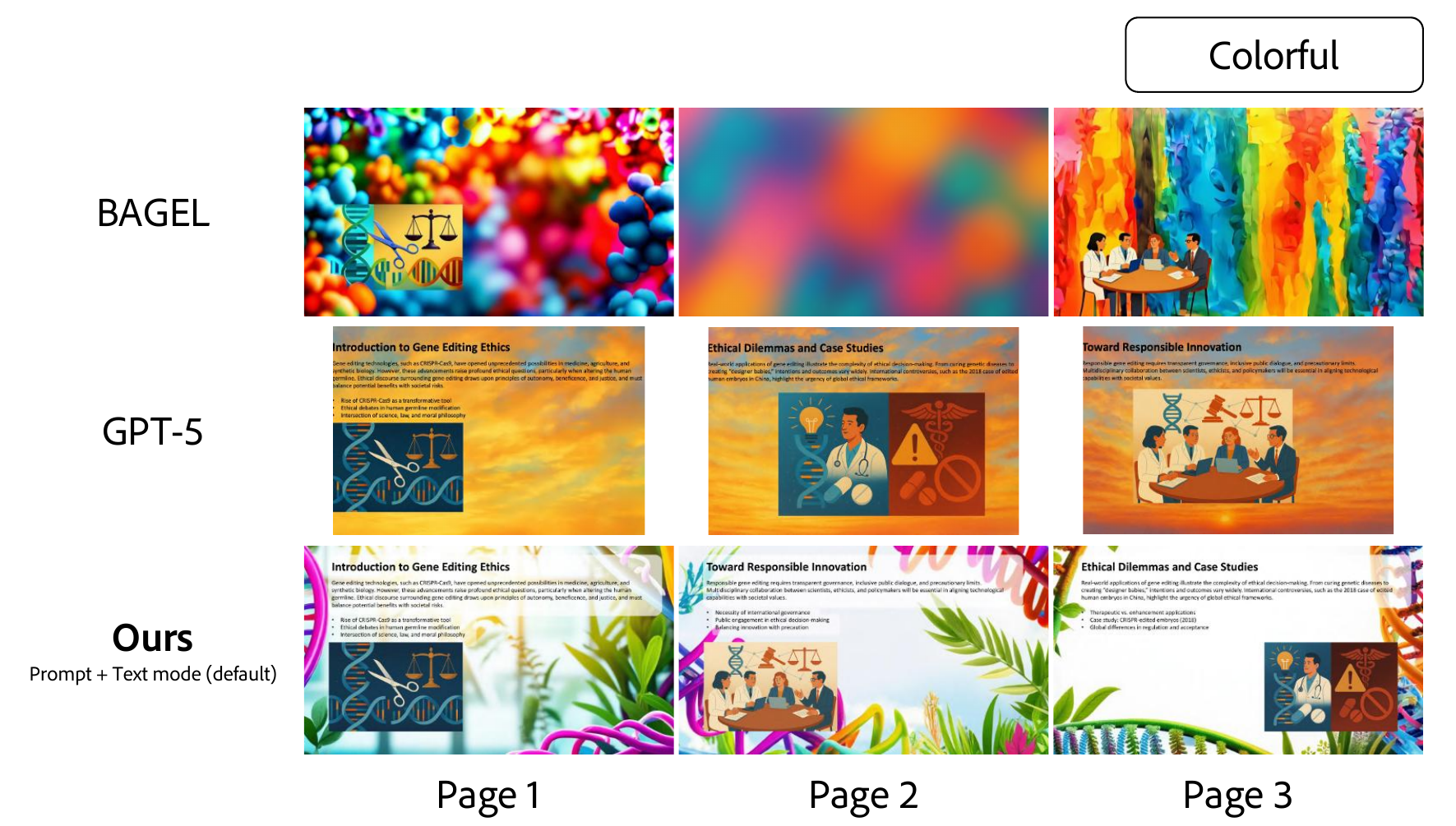}
    \caption{Comparison of background generation under the \emph{Colorful} style slides.}
\end{figure}

\begin{figure}[t]
    \centering
    \includegraphics[width=\linewidth]{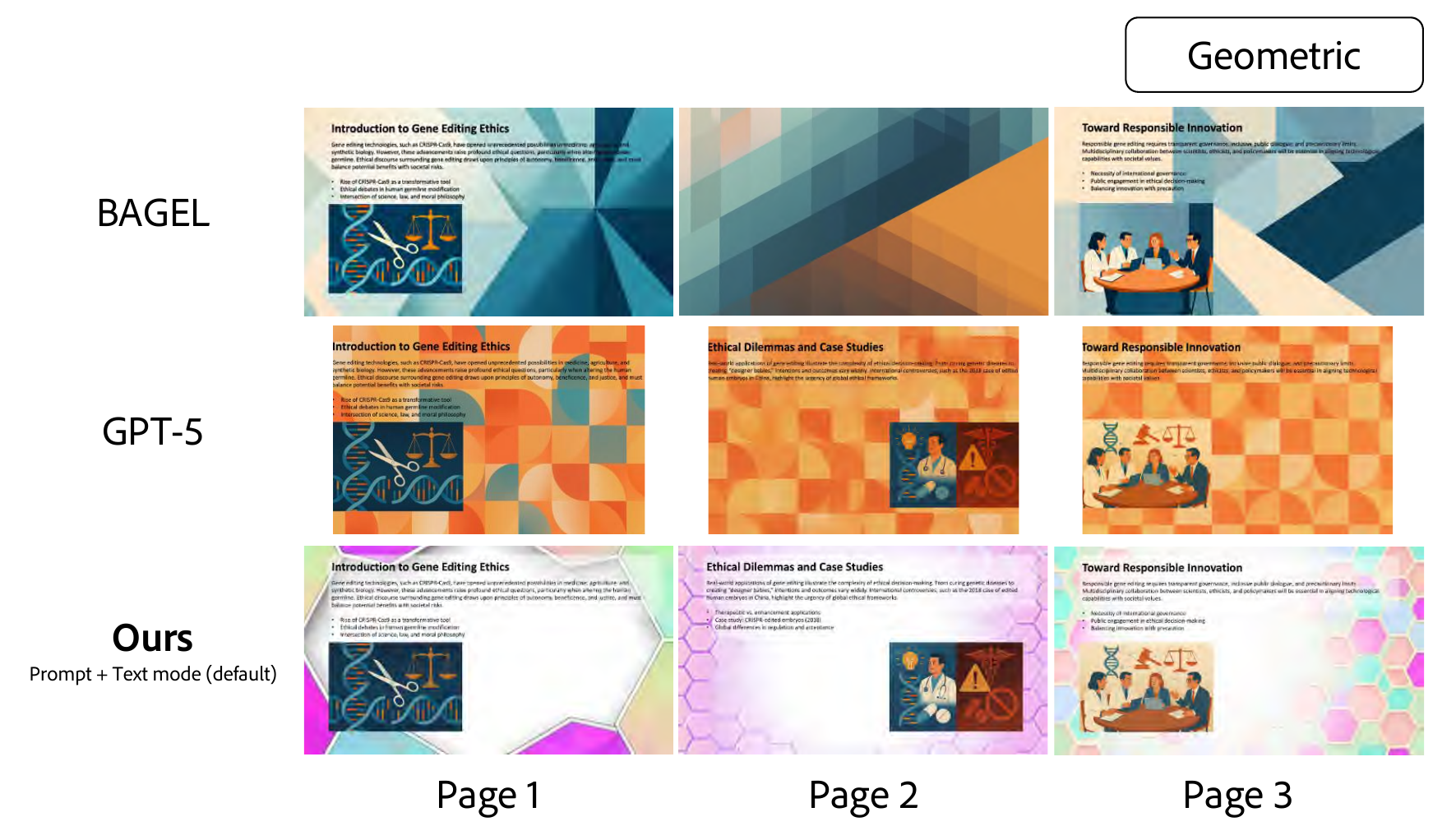}
    \caption{Comparison of background generation under the \emph{Geometric} style slides.}
\end{figure}

\newpage

\begin{figure}[t]
    \centering
    \includegraphics[width=\linewidth]{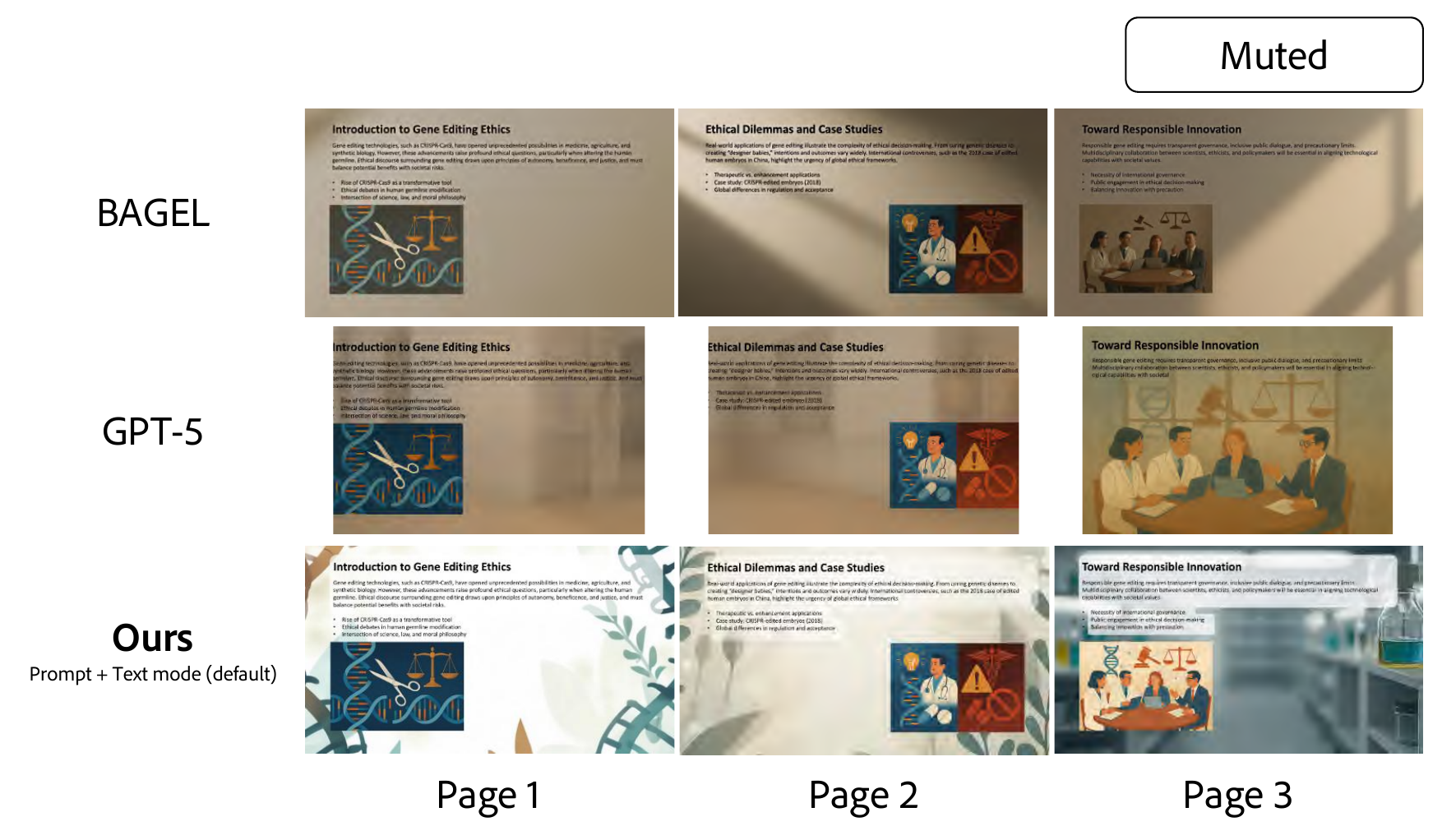}
    \caption{Comparison of background generation under the \emph{Muted} style slides.}
\end{figure}

\begin{figure}[t]
    \centering
    \includegraphics[width=\linewidth]{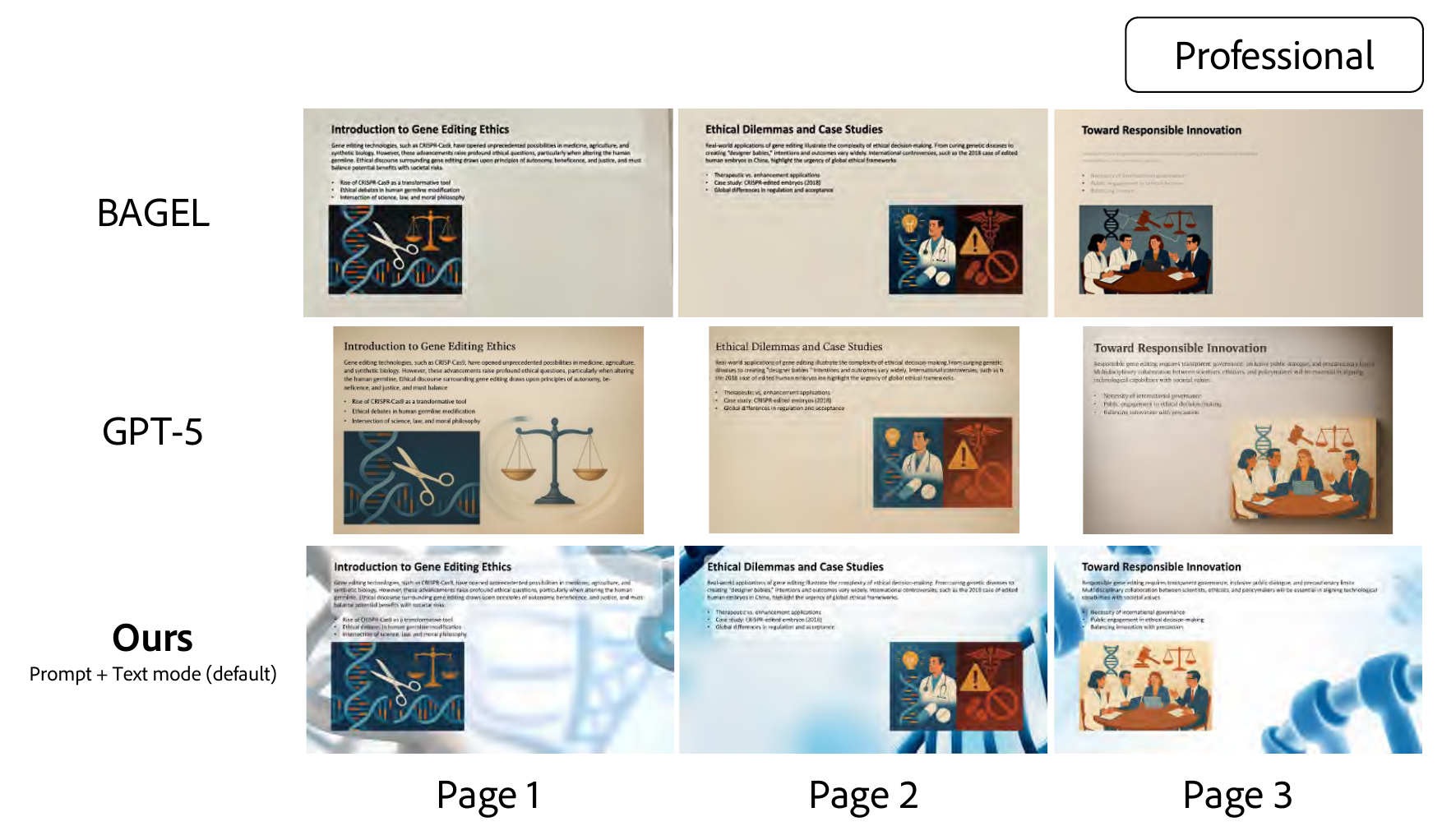}
    \caption{Comparison of background generation under the \emph{Professional} style slides.}
\end{figure}

\newpage

\begin{figure}[t]
    \centering
    \includegraphics[width=\linewidth]{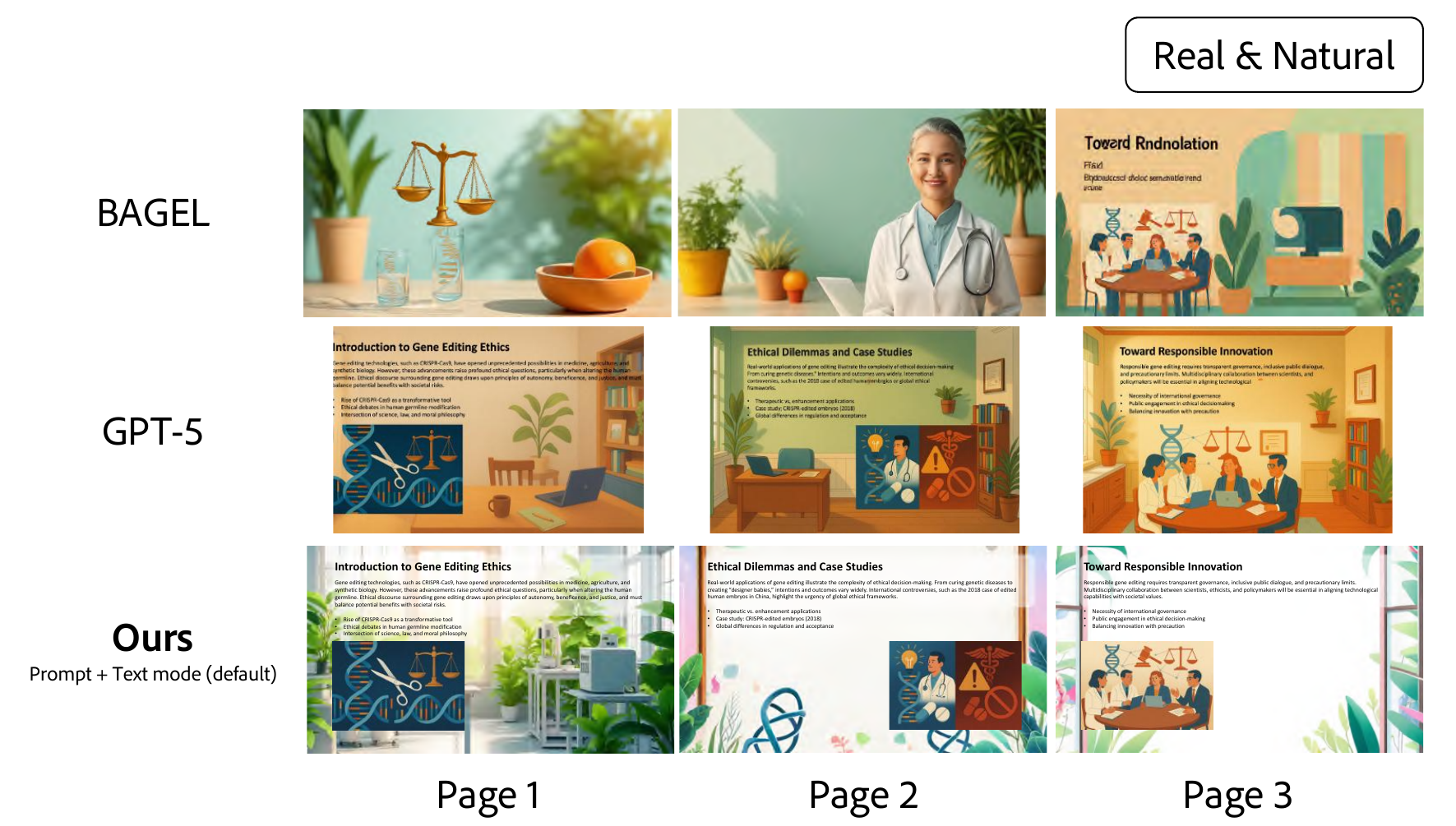}
    \caption{Comparison of background generation under the \emph{Real \& Natural} style slides.}
\end{figure}

\begin{figure}[t]
    \centering
    \includegraphics[width=\linewidth]{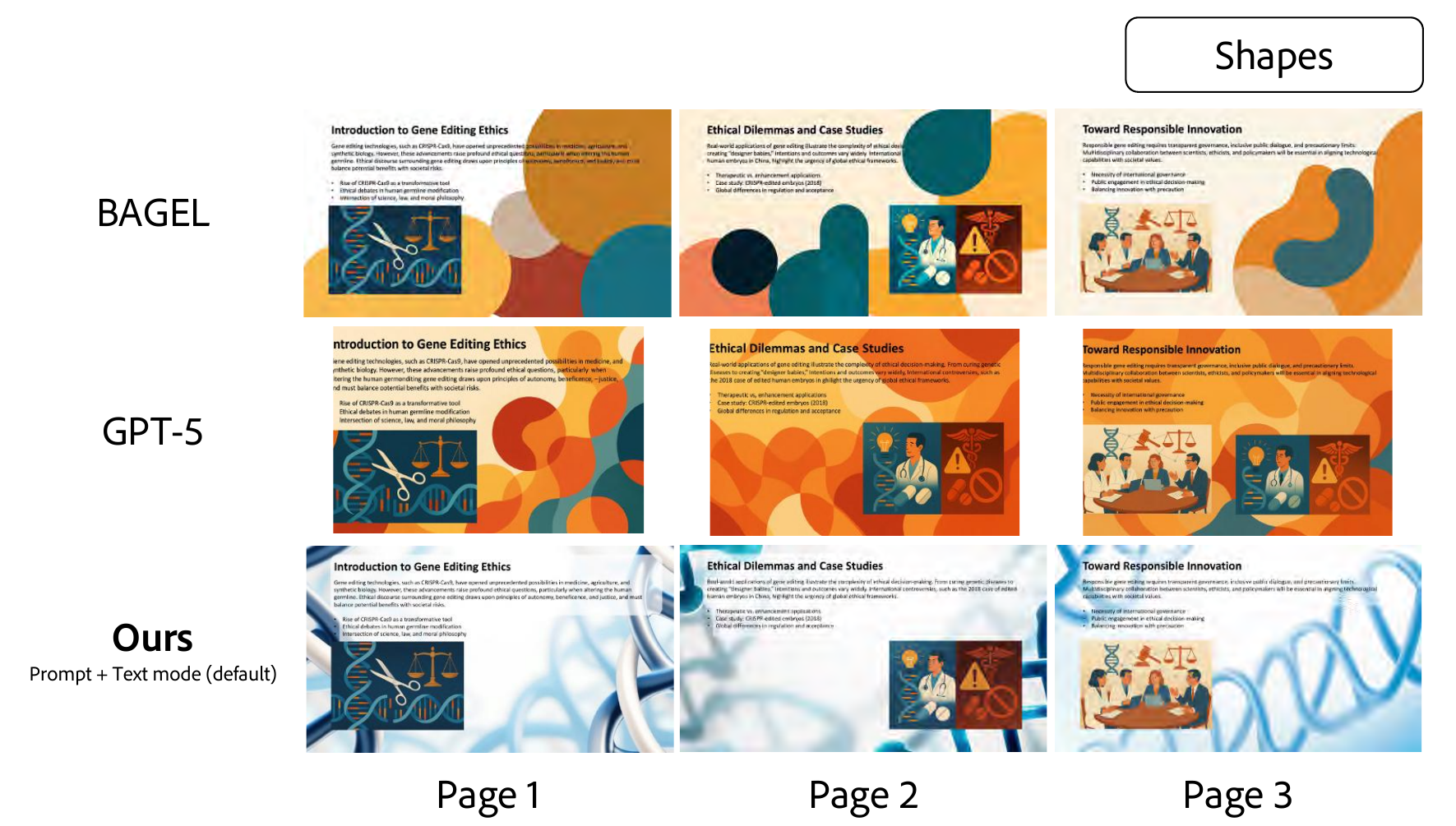}
    \caption{Comparison of background generation under the \emph{Shapes} style slides.}
\end{figure}

\newpage

\begin{figure}[t]
    \centering
    \includegraphics[width=\linewidth]{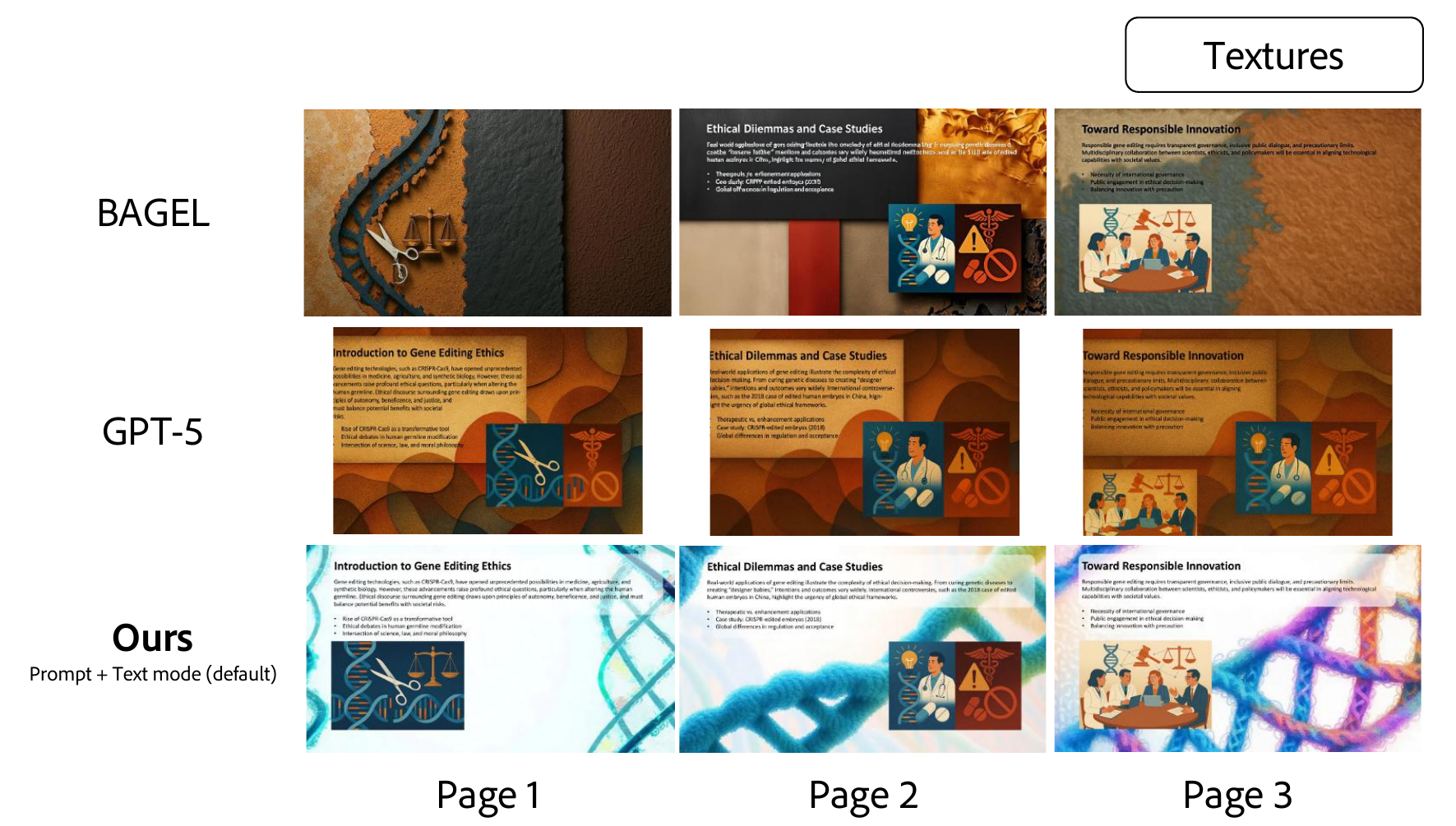}
    \caption{Comparison of background generation under the \emph{Textures} style slides.}
\end{figure}

\newpage

\begin{figure}[t]
    \centering
    \includegraphics[width=\linewidth]{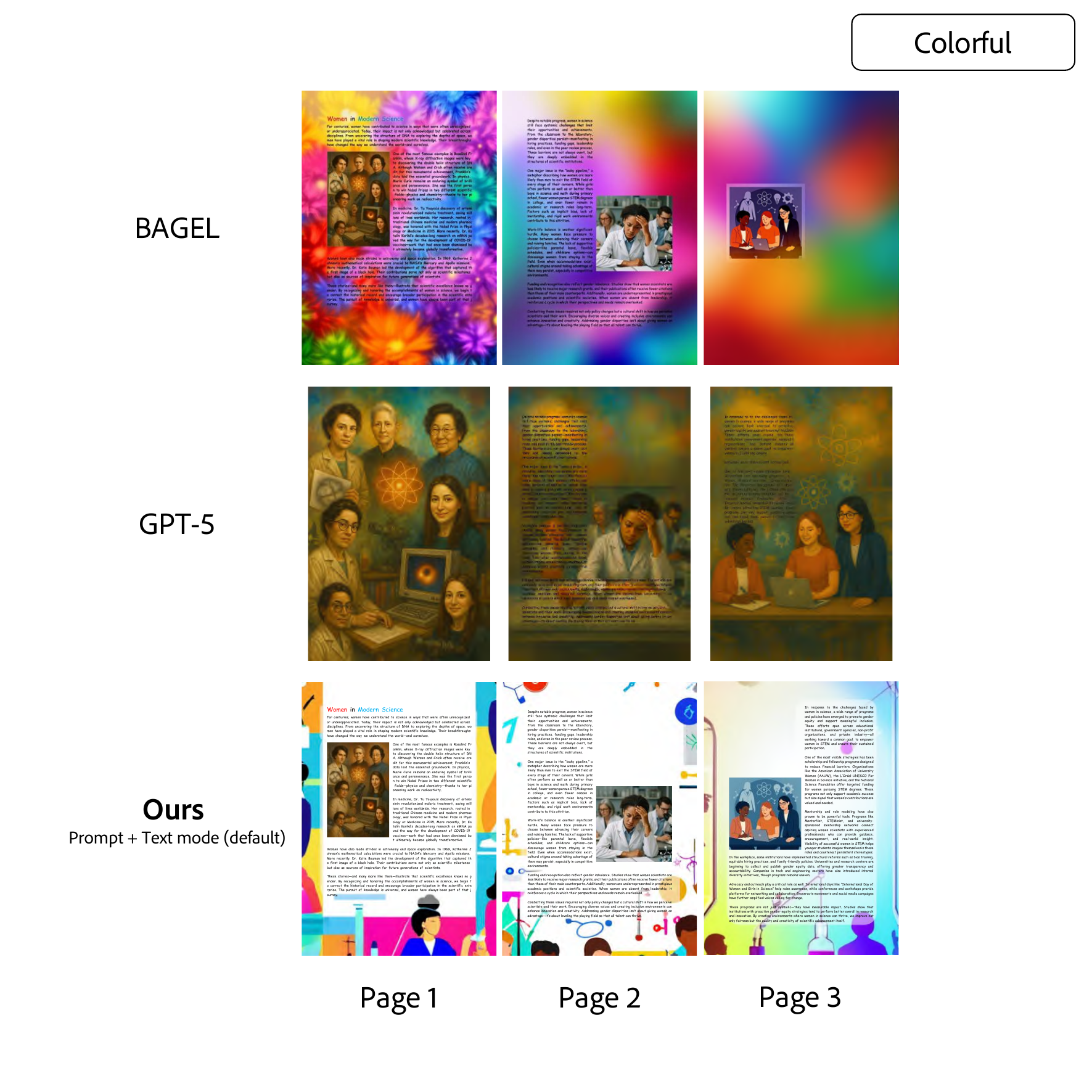}
    \caption{Comparison of background generation under the \emph{Colorful} style PDFs.}
\end{figure}

\newpage

\begin{figure}[t]
    \centering
    \includegraphics[width=\linewidth]{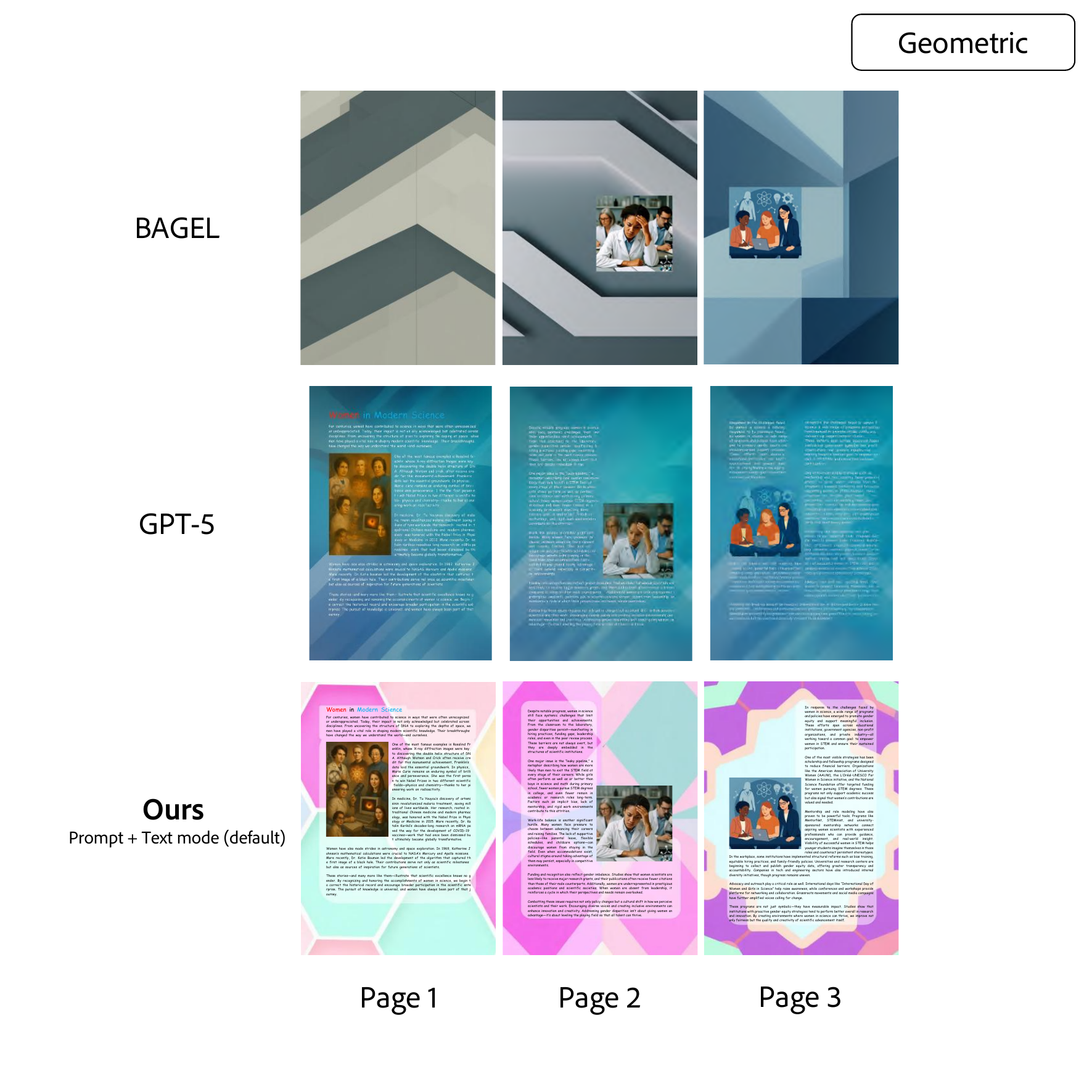}
    \caption{Comparison of background generation under the \emph{Geometric} style PDFs.}
\end{figure}

\newpage

\begin{figure}[t]
    \centering
    \includegraphics[width=\linewidth]{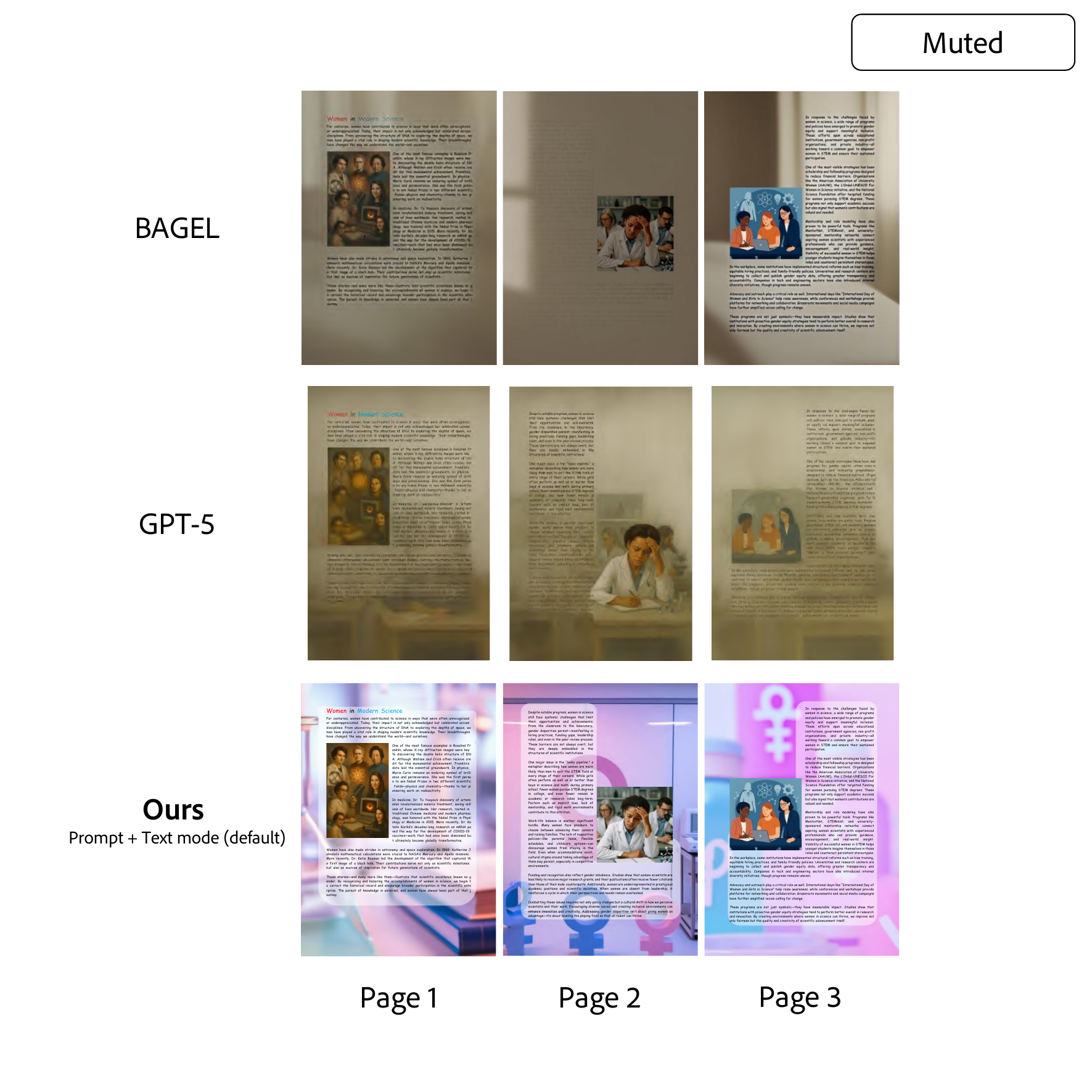}
    \caption{Comparison of background generation under the \emph{Muted} style PDFs.}
\end{figure}

\newpage

\begin{figure}[t]
    \centering
    \includegraphics[width=\linewidth]{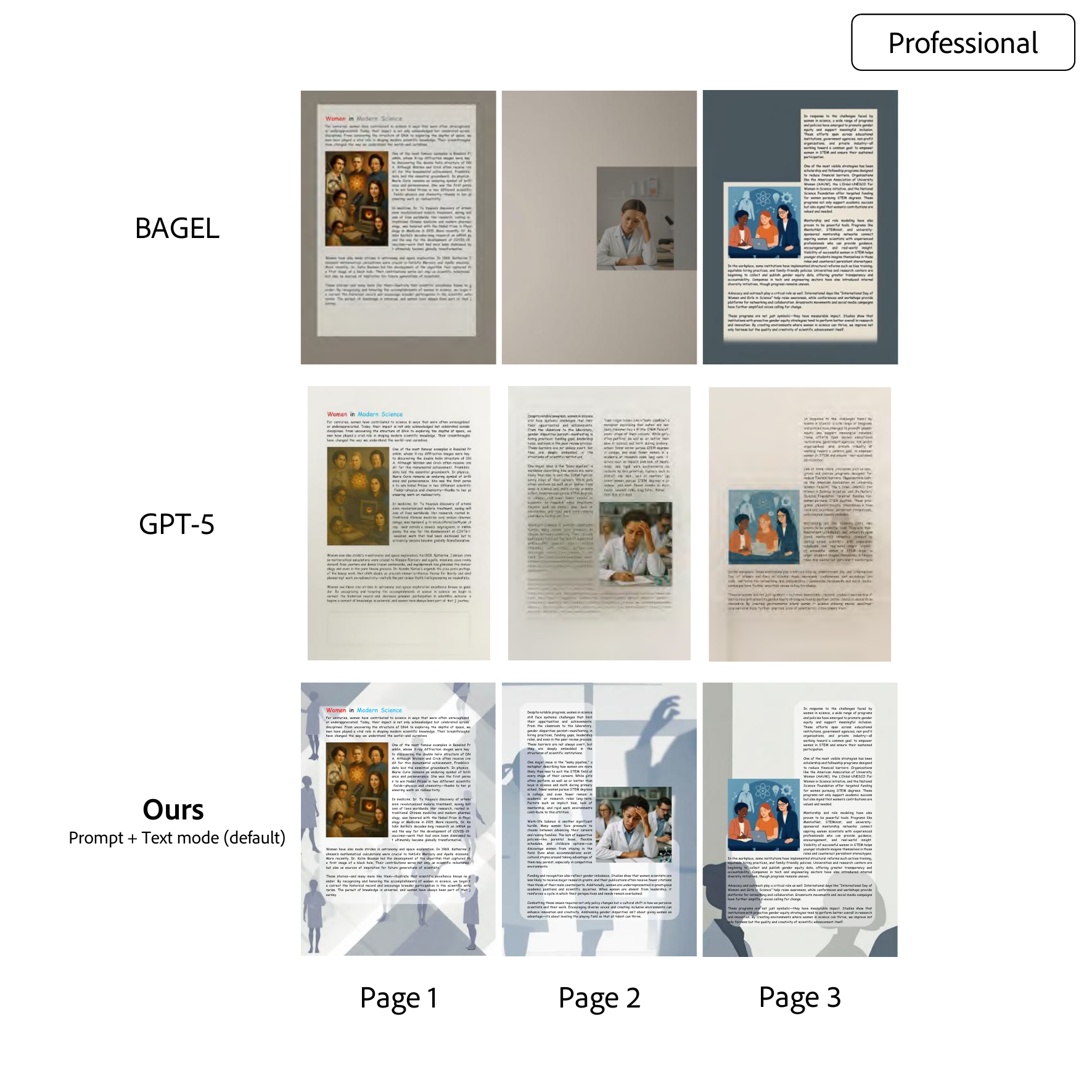}
    \caption{Comparison of background generation under the \emph{Professional} style PDFs.}
\end{figure}

\newpage

\begin{figure}[t]
    \centering
    \includegraphics[width=\linewidth]{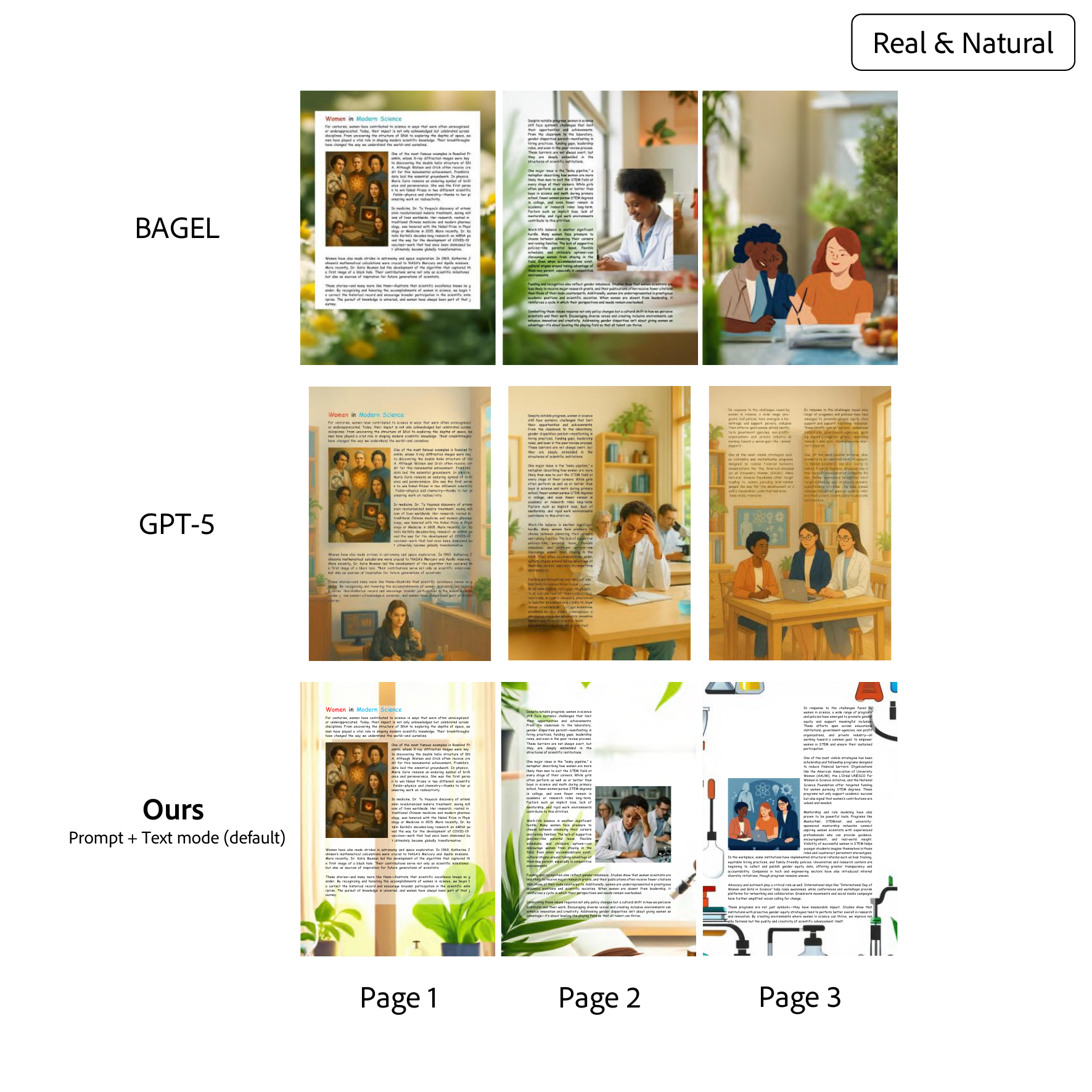}
    \caption{Comparison of background generation under the \emph{Real \& Natural} style PDFs.}
\end{figure}

\newpage

\begin{figure}[t]
    \centering
    \includegraphics[width=\linewidth]{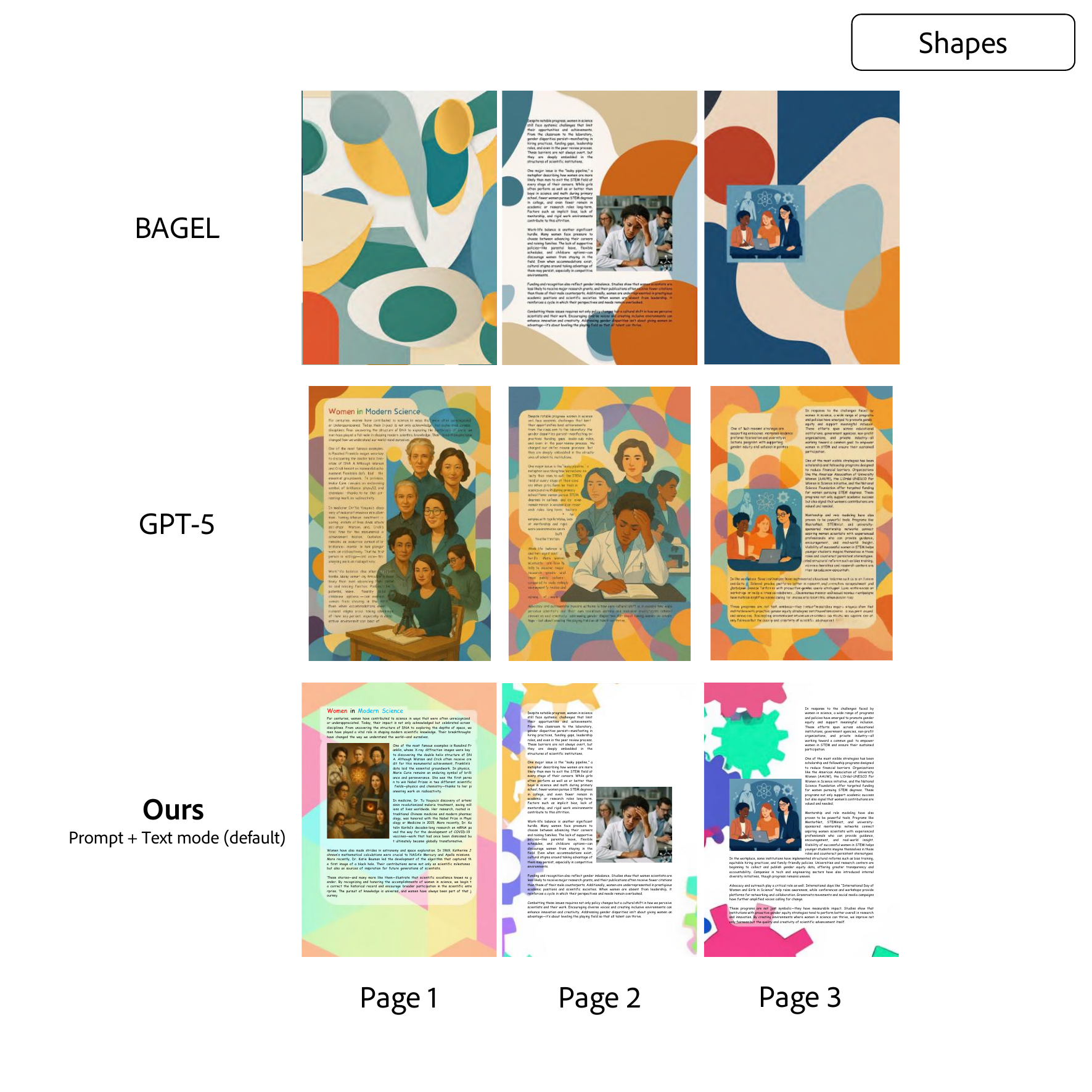}
    \caption{Comparison of background generation under the \emph{Shapes} style PDFs.}
\end{figure}

\newpage

\begin{figure}[t]
    \centering
    \includegraphics[width=\linewidth]{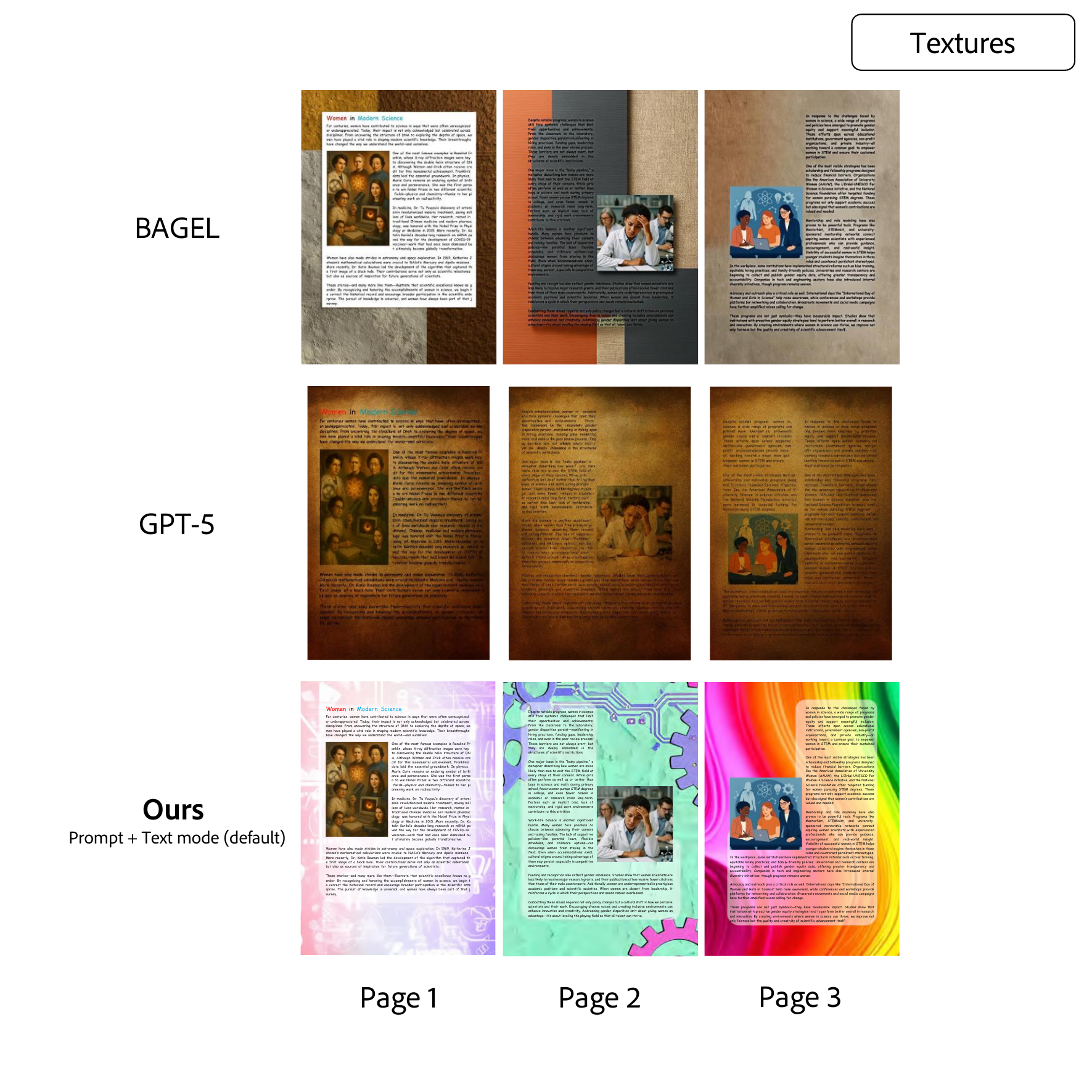}
    \caption{Comparison of background generation under the \emph{Textures} style PDFs.}
\end{figure}

\newpage

\begin{figure}[t]
    \centering
    \includegraphics[width=\linewidth]{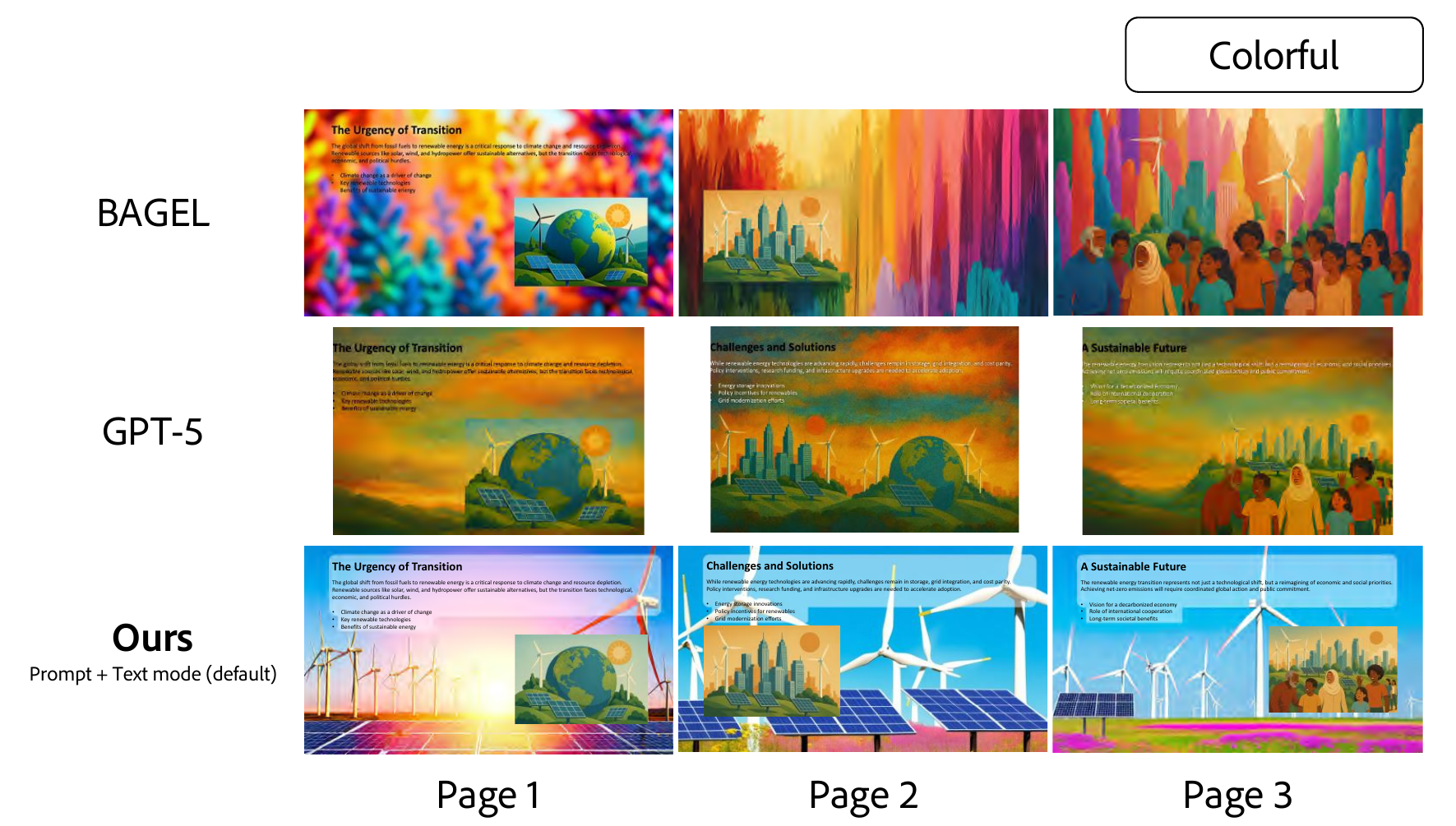}
    \caption{Comparison of background generation under the \emph{Colorful} style slides.}
\end{figure}

\begin{figure}[t]
    \centering
    \includegraphics[width=\linewidth]{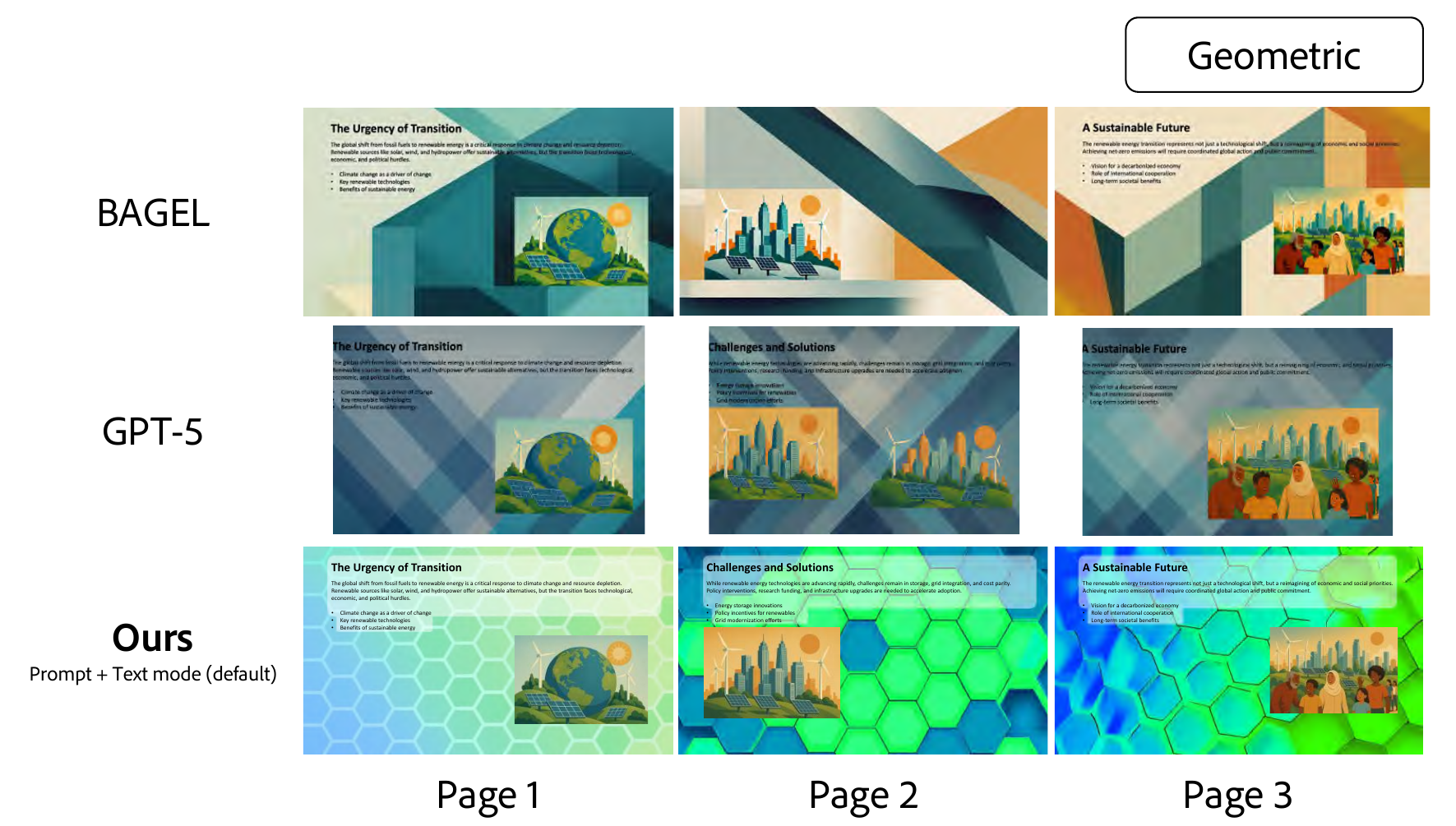}
    \caption{Comparison of background generation under the \emph{Geometric} style slides.}
\end{figure}

\newpage

\begin{figure}[t]
    \centering
    \includegraphics[width=\linewidth]{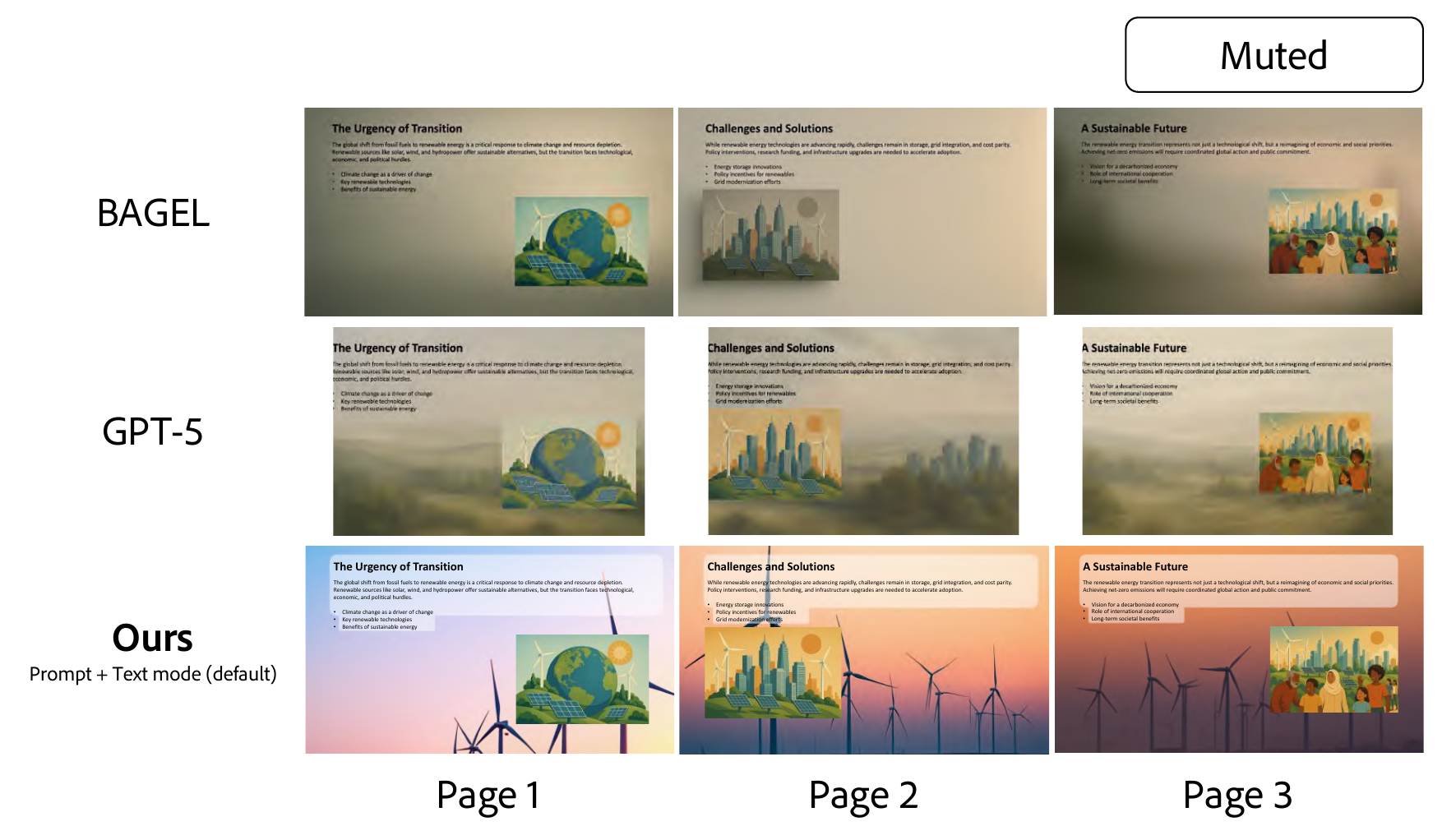}
    \caption{Comparison of background generation under the \emph{Muted} style slides.}
\end{figure}

\begin{figure}[t]
    \centering
    \includegraphics[width=\linewidth]{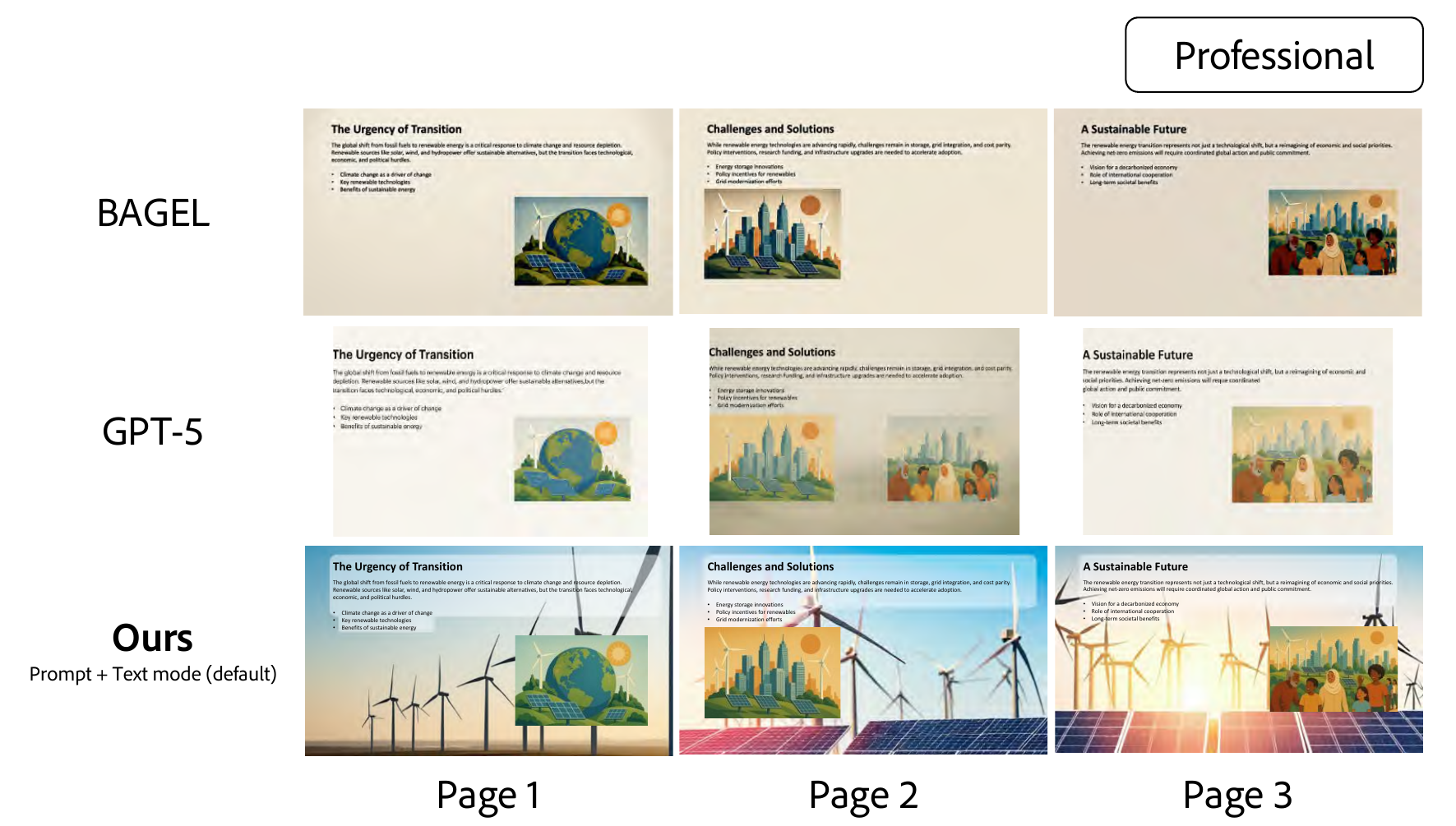}
    \caption{Comparison of background generation under the \emph{Professional} style slides.}
\end{figure}

\newpage

\begin{figure}[t]
    \centering
    \includegraphics[width=\linewidth]{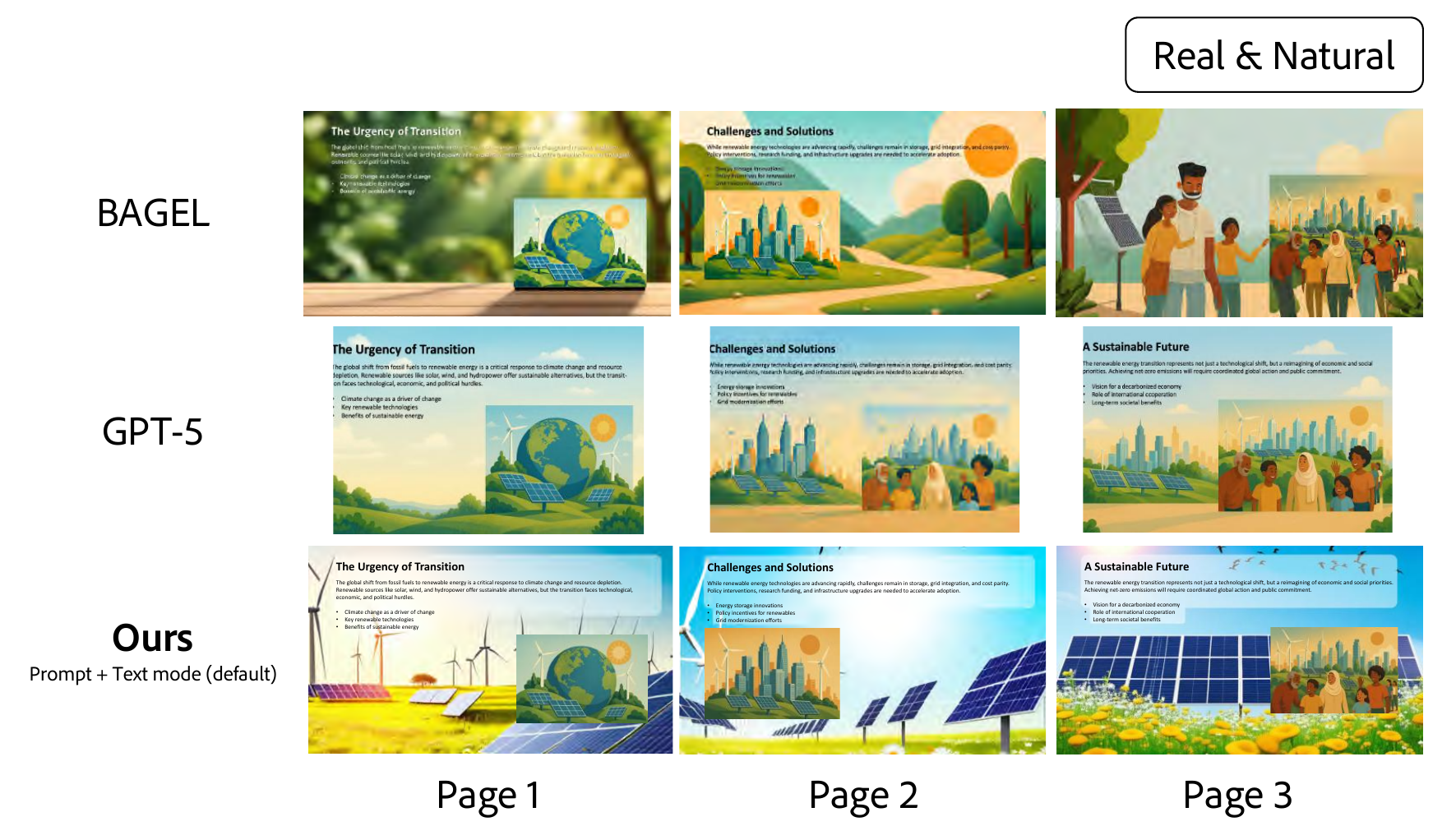}
    \caption{Comparison of background generation under the \emph{Real \& Natural} style slides.}
\end{figure}

\begin{figure}[t]
    \centering
    \includegraphics[width=\linewidth]{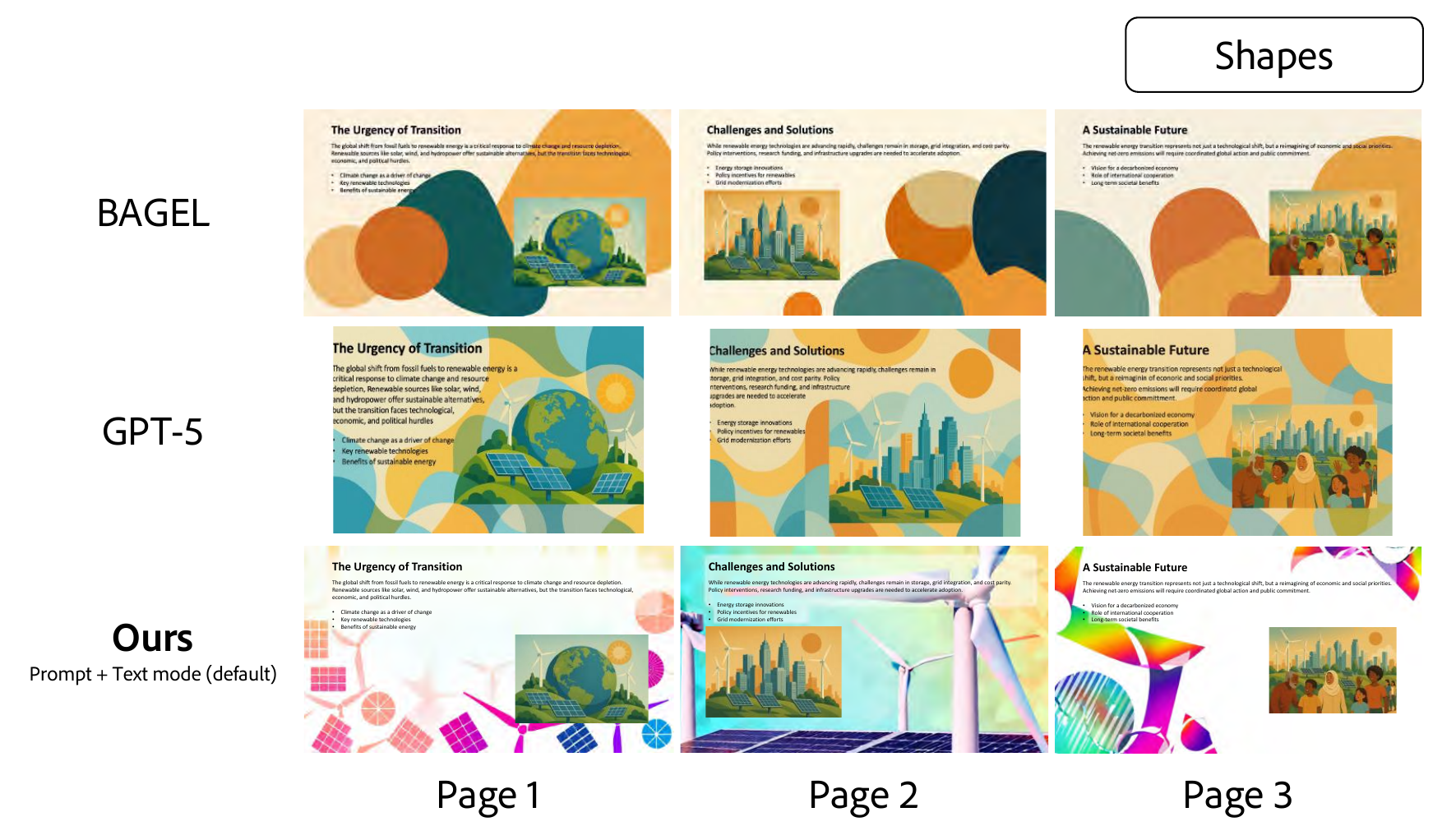}
    \caption{Comparison of background generation under the \emph{Shapes} style slides.}
\end{figure}

\newpage

\begin{figure}[t]
    \centering
    \includegraphics[width=\linewidth]{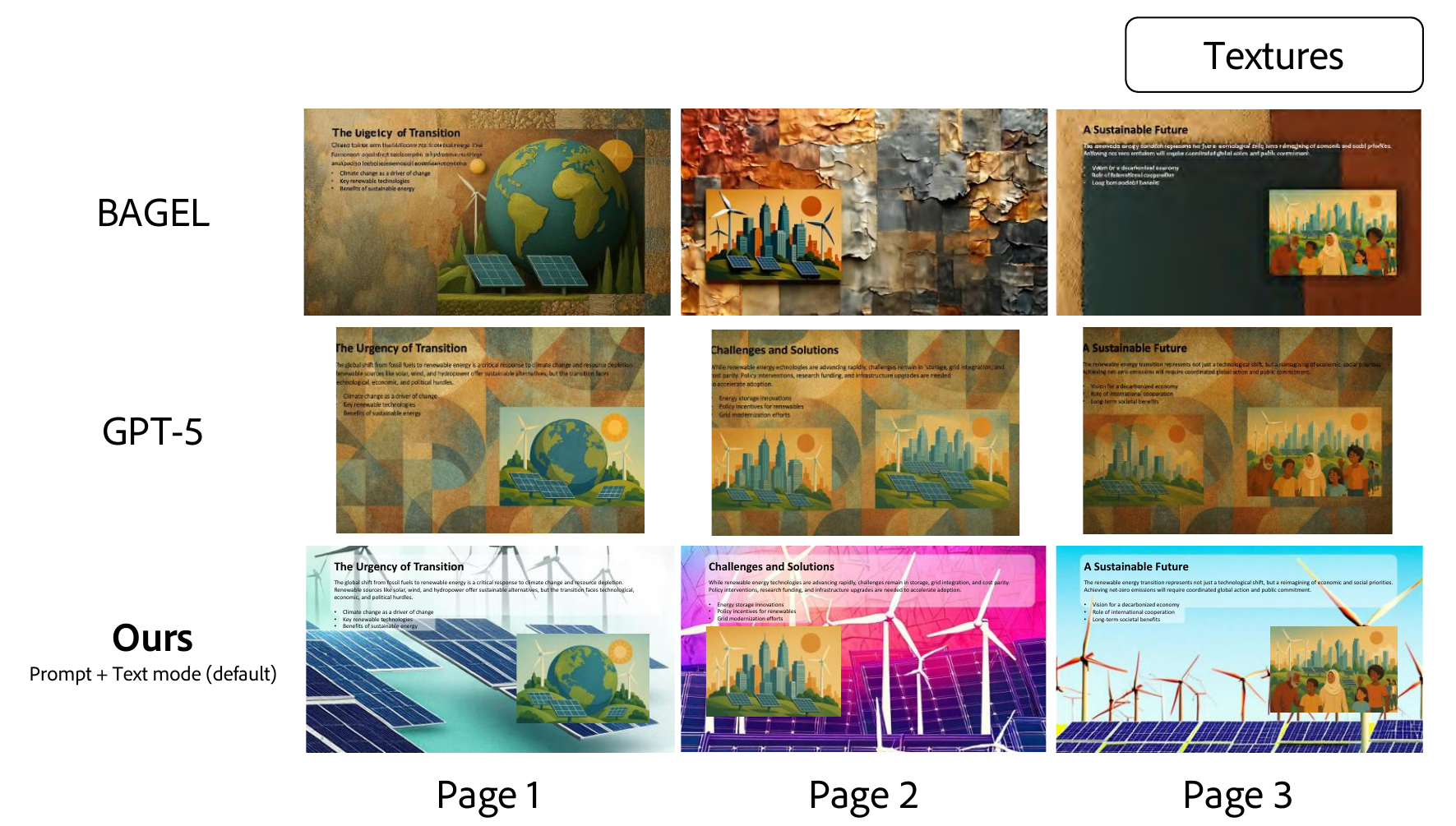}
    \caption{Comparison of background generation under the \emph{Textures} style slides.}
\end{figure}

\end{document}